\documentclass[10pt,twocolumn,letterpaper]{article}

\usepackage[pagenumbers]{iccv} 

\definecolor{iccvblue}{rgb}{0.21,0.49,0.74}
\usepackage[pagebackref,breaklinks,colorlinks,allcolors=iccvblue]{hyperref}
\usepackage{amsmath}
\usepackage{tabularx}
\usepackage{hyperref}
\usepackage{subcaption}
\usepackage{graphicx}
\usepackage[accsupp]{axessibility}
\usepackage[capitalize]{cleveref}
\newcolumntype{Y}{>{\hsize=1.1\hsize}X}  
\newcolumntype{Z}{>{\hsize=0.8\hsize}X}  

\def\paperID{9021} 
\def\confName{ICCV}
\def\confYear{2025}

\title{HumorDB: Can AI understand graphical humor?}

\author{Vedaant V Jain\\
University of Illinois Urbana-Champaign\\
{\tt\small vvjain3@illinois.edu}
\and
Felipe dos Santos Alves Feitosa\\
University of São Paulo\\
{\tt\small felipefeitosa@usp.br}
\and
Gabriel Kreiman\\
Harvard Medical School\\
{\tt\small gabriel.kreiman@tch.harvard.edu}
}

\begin{document}
\maketitle
\begin{abstract}
Despite significant advancements in image segmentation and object detection, understanding complex scenes remains a significant challenge. Here, we focus on graphical humor as a paradigmatic example of image interpretation that requires elucidating the interaction of different scene elements in the context of prior cognitive knowledge. This paper introduces \textbf{HumorDB}, a novel, controlled, and carefully curated dataset designed to evaluate and advance visual humor understanding by AI systems. The dataset comprises diverse images spanning photos, cartoons, sketches, and AI-generated content, including minimally contrastive pairs where subtle edits differentiate between humorous and non-humorous versions. We evaluate humans, state-of-the-art vision models, and large vision-language models on three tasks: binary humor classification, funniness rating prediction, and pairwise humor comparison. The results reveal a gap between current AI systems and human-level humor understanding. While pretrained vision-language models perform better than vision-only models, they still struggle with abstract sketches and subtle humor cues. Analysis of attention maps shows that even when models correctly classify humorous images, they often fail to focus on the precise regions that make the image funny. Preliminary mechanistic interpretability studies and evaluation of model explanations provide initial insights into how different architectures process humor. Our results identify promising trends and current limitations, suggesting that an effective understanding of visual humor requires sophisticated architectures capable of detecting subtle contextual features and bridging the gap between visual perception and abstract reasoning.
All the code and data are available here: 
\href{https://github.com/kreimanlab/HumorDB}{https://github.com/kreimanlab/HumorDB}
\end{abstract}
\footnote{$20^{th}$ International Conference on Computer Vision, ICCV 2025 (Poster)}    
\section{Introduction}
\label{sec:intro}

The last decade has seen remarkable strides in computer vision, enabling systems to segment images, label objects, and even write sophisticated captions. Despite these successes, the problem of scene understanding remains challenging. Interpreting a scene often requires elucidating the relationship between objects and their positions, the intention of agents, and linking visual information with prior knowledge. This challenge is particularly evident in tasks that require higher-level cognitive processes, such as recognizing and interpreting humor in visual content. Graphical humor understanding demands a high level of cognitive abstraction, as it requires context awareness, expectations, cultural knowledge, and the identification of incongruities \cite{humor_reasonincongru}. Thus, graphical humor serves as an ideal testbed for scene understanding capabilities. 

\begin{figure}[t]
\centering
\includegraphics[width=0.8\linewidth]{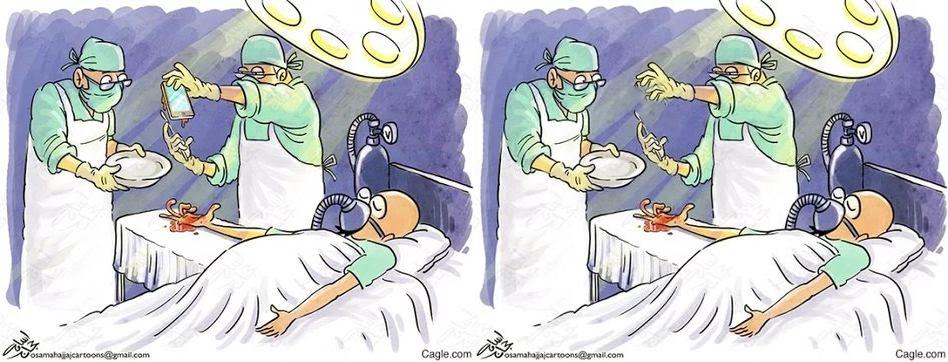}
\caption{\textbf{Example image pair}. Left: image rated as funny (83.3\% of participants). Right: modified image rated as not funny (85.7\% of participants). Focus on the phone in the surgeon's hand in the left image.}
\label{fig:image_pair}
\end{figure}

Consider \textbf{Fig. \ref{fig:image_pair}}, left. To comprehend what is going on, the viewer needs to detect locations (surgical setting), agents (patient, medical practitioners), and objects (cell phone, hand). The grasping in the hand induces us to think that the medical experts have ``surgically excised'' the phone from the hand. Given the prominent role of cell phones, many people have jokingly stated that phones are physically attached to hands, and the image plays with this idea. Of note, the reader has likely never seen this particular image or any similar image before. Upon first exposure to this image, readers can rapidly interpret what is going on and may consider the image to be somewhat humorous (83.3\% of participants indicated that this image is funny; more on how this number was computed is described in \textbf{Sec.~\ref{sec:data_qual_part_rel}}). In stark contrast, consider \textbf{Fig. \ref{fig:image_pair}}, right. The two images are identical except that the cell phone was removed. Despite the strong similarity between the two images, the one on the right is no longer humorous (85.7\% of participants indicated that the image on the right is \emph{not} funny).

Computational humor understanding extends beyond technical complexity to practical implications for human-AI interaction, content moderation, and creative applications. However, existing datasets often lack rigorous controls, exhibit limited diversity, or focus primarily on text rather than visual humor.  To address this, we introduce \textbf{HumorDB}, a dataset of images including photos, sketches and AI-generated content. Drawing inspiration from action recognition research (e.g., \cite{Jacquot_2020_CVPR}), a key innovation in the dataset is the inclusion of pairs of minimally different images with contrasting humor ratings
distinguished only by the humorous element like the one in \textbf{Fig. \ref{fig:image_pair}}, enabling precise evaluation of models' ability to detect humor-inducing elements while controlling for confounding factors. We also include a benchmark consisting of quantitative human measurements in three different tasks using these images: Binary classification (funny vs. not funny), Range rating (scoring images on a scale of 1-10), and Comparison (deciding which of two images is funnier). 
 We compare human responses with both vision-only models (e.g., \cite{he2015deep, liu2022convnet, oquab2023dinov2, dehghani2023scaling}) and vision-language models (e.g., \cite{li2022blip, liu2023llava, liu2023improvedllava, openai2023gpt4, geminiteam2024gemini}) across diverse image types and analyze models' explanations using mechanistic interpretability techniques. 
 HumorDB enables precise assessment of models’ ability to detect humor-inducing features.

Our findings reveal a substantial gap between current models and human-level humor understanding. State-of-the-art models outperform chance but fall below humans.
The comparison task proves particularly challenging, with models showing limited ability to make nuanced judgments between similar images. Notably, large multimodal language models demonstrate robust zero-shot performance, suggesting promising directions for bridging this gap. Beyond performance evaluation, we leverage HumorDB's contrastive pairs to analyze how humor understanding emerges in neural networks through attention map analysis and mechanistic interpretability techniques.

\section{Related Work}
The best-known theory of humor is probably the ``incongruity theory'' which posits that humor arises from unexpected subversions of contextual expectations \cite{humor_incongruity}. 
Recent datasets for computational humor understanding \cite{UR_Funny, humor_hindi} include multi-modal humor detection \cite{UR_Funny, humor_hindi, multi_modalkan}, funny image captioning \cite{Humor_captions}, abstract clipart scenes with object replacements \cite{Chandrasekaran_2016_CVPR}, and matching captions to cartoons \cite{new_yorkercaptions}. However, most efforts have focused on either textual humor \cite{humor_text, humor_text1, humor_text2, humor_socialmedia} or multimodal contexts \cite{humor_video, humor_video1, multi_modalkan} rather than purely visual humor and gaining contextual clues may be easier to get from one modality than another. 

Computational approaches to humor understanding have involved using attention-based Bi-LSTM for identifying humor in text on social media \cite{humor_socialmedia}, generation of negative examples \cite{dutch_genexamp}, the use of transformers \cite{humor_multimodaltransf}, and CRF-RNN-CNN \cite{humor_audio}. However, most of these approaches have focused on either natural language \cite{humor_text, humor_text1, humor_text2, humor_socialmedia} or multimodal contexts (especially videos \cite{humor_video, humor_video1, multi_modalkan}).

HumorDB focuses on vision and significantly extends previous datasets via (1) diversity of realistic images (photos, cartoons, sketches, AI-generated), (2) sophisticated controls to distill humor from confounding factors, and (3) a comprehensive evaluation framework including human metrics across three different tasks.


\section{Methods}
\label{sec:formatting}

All the code and data are available here: 
\href{https://github.com/kreimanlab/HumorDB}{https://github.com/kreimanlab/HumorDB}

\subsection{Building HumorDB}
We constructed a diverse repository of images through (i) web scraping via the Google Search API with broad queries and specific prompts followed by manual curation, and (ii) generating novel images using AI-based models like DALL-E \cite{dalle2_openai} and MidJourney \cite{midjourney}. We capped each search result to 15 images to avoid style biases. The resulting images consisted of: Photos (36\%), Photoshopped real-life photos (35\%), Cartoons (14\%), Sketches (5\%), and AI-Generated images (10\%). We removed images with explicit/offensive material. We removed cases where humor relied on embedded text but kept any non-humor-related text (e.g., artist names) to retain authenticity in the images.

A central innovation is our collection of paired images whose only differences involve removing or modifying the humorous element (e.g., \textbf{Fig. \ref{fig:image_pair}}). This was achieved by either 
(i) Photo Editing, i.e., using Adobe Photoshop to carefully remove, obscure, or replace objects that contributed to humor, or (ii) AI Inpainting, i.e., employing stable diffusion-based inpainting for more naturalistic edits that seamlessly replace incongruous elements while preserving the rest of the image.
Through this process, we generated 1,271 pairs of original (``funny'') and modified (``not funny'') counterparts,ensuring visual similarity except for key comedic features. Importantly, to mitigate potential editing artifacts, we applied \textbf{subtle enhancements} to \textbf{both versions (funny and not funny)} of each pair. We further validated that models learn humor-relevant features, not just editing artifacts, by also testing on a control set of 650 non-modified, non-funny images (see Sec.~\ref{sec:results} for results and Appendix~\ref{sec:expts} for details).
Throughout the manuscript, ``original images'' refer to humor-laden content, while ``modified images'' are those we altered to remove the humor.

\subsection{Assessing human performance}
\label{sec:data_qual_part_rel}
We conducted online psychophysics experiments with 650 participants through Amazon Mechanical Turk and Prolific to gather human evaluations. Each participant rated 100 images in one of three tasks:
\begin{enumerate}
\item \textbf{Binary Task:} Two-alternative forced choice classification as ``Funny'' or ``Not Funny''
\item \textbf{Range Task:} Rating funniness on a scale of 1 (Not at all funny) to 10 (Extremely funny)
\item \textbf{Comparison Task:} Selecting the funnier of two side-by-side images, followed by providing a brief justification for the choice
\end{enumerate}

\noindent \textbf{Quality control}. We implemented 5 controls:
\begin{enumerate}
\item \textbf{Minimum Viewing Time:} 500ms minimum viewing time per image before response submission (based on \cite{dreamsimlearningnewdimensions}), with task termination after 4 violations
\item \textbf{Repeated Images:} 10\% of images were  repeated at random time points to assess self-consistency
\item \textbf{Outlier Removal:} Participants with $>$10\% of ratings having z-scores outside $\pm$1.96 were excluded
\item \textbf{Textual Justifications:} Brief justification of choices verified attention to content. Only in comparison task.
\item \textbf{Unique Task Enrollment:} Prevention of repeated participation in the same task
\end{enumerate}

\noindent \textbf{Comparison Task Design}.
To make pairwise comparisons tractable, we employed a stratified sampling approach:
\begin{enumerate}
\item Divided images into 8 strata using mean Binary ratings
\item Selected five representative images per stratum (40 total)
\item Collected additional Binary ratings for validation
\item Identified an anchor image per stratum for comparisons. All comparisons had at least one of the 8 anchor images.
\end{enumerate}

\noindent \textbf{Participant Reliability Assessment}.
For the repeated images (10\%), we applied the following exclusion criteria:
\begin{itemize}
\item Binary and Comparison Tasks: Excluded participants with $>$3 inconsistent responses on repeated images
\item Range Task: Excluded participants if any single image received ratings differing by $>$4 points, or if $>$4 images showed differences exceeding 2 points
\end{itemize}

\begin{figure}
\centering
\includegraphics[height=0.5\linewidth]{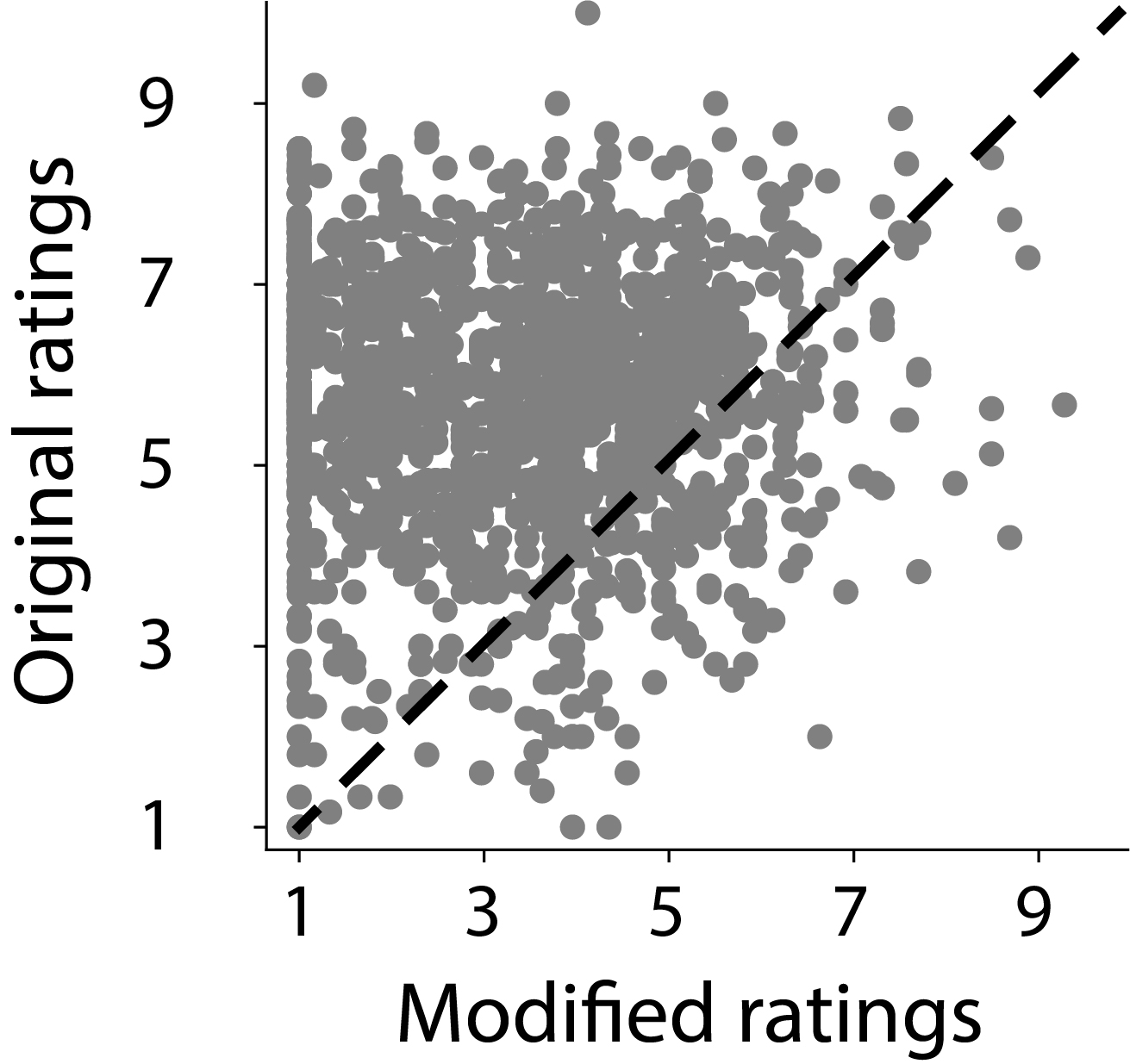}
\caption{\textbf{Modifications rendered images less humorous}. Each point compares the rating of image pairs (y-axis: original, x-axis: modified pair; total 1,271 pairs; line = identity). For the majority of images ($86.4 \%$), the ratings for the original images were higher. 
}
\label{fig:original_modified_ratings}
\end{figure}

\begin{figure}
\centering
\includegraphics[height=0.5\linewidth]{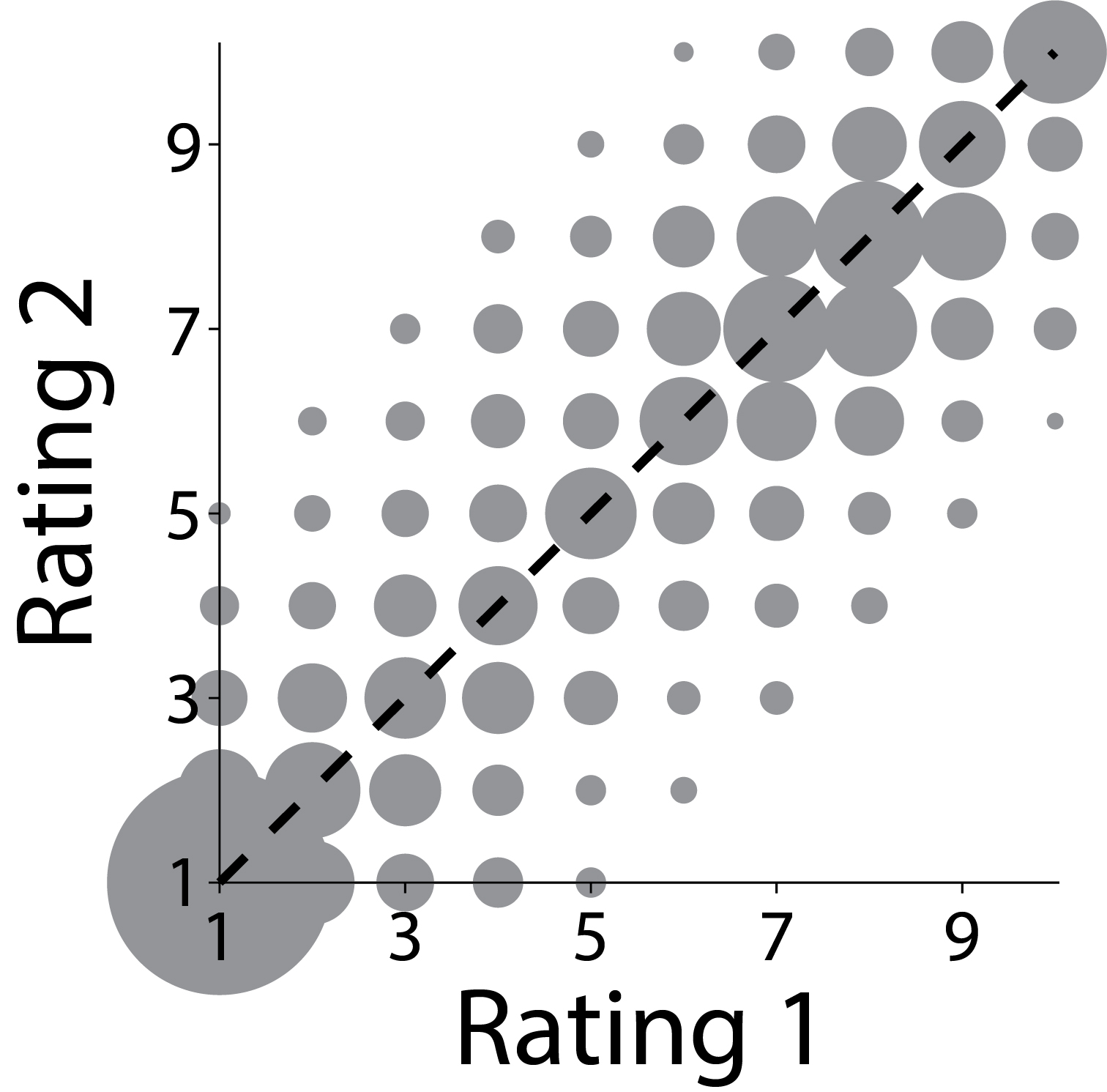}
\caption{\textbf{Participants showed high self-reliability} (Range Ratings). Higher ratings denote more humorous images.
For each participant, 10 images were presented twice at random time points to assess reliability. There was a strong correlation ($\rho$ = 0.89) between the first and second ratings (1,800 pairs; circle sizes indicate number of ratings/pair). The dashed line shows the diagonal.}
\label{fig:user_ratings_compare}
\end{figure}

The results showed strong self-reliability in all tasks:
\begin{itemize}
\item {Binary Task:} The degree of self-consistency for repeated images was $84.2\pm13.3\%$
\item {Range Task:} Numerical judgments were highly stable ($\rho = 0.89$, Fig.\ref{fig:user_ratings_compare})
\item {Comparison Task:} The degree of self-consistency for repeated choices was $91.3\pm14.8\%$ 
\item {Original vs. Modified:} Original images were consistently rated as funnier than their modified counterparts (86.4\% of pairs, \textbf{Fig.\ref{fig:original_modified_ratings}})
\end{itemize}

\begin{figure}
\centering
\includegraphics[height=0.5\linewidth]{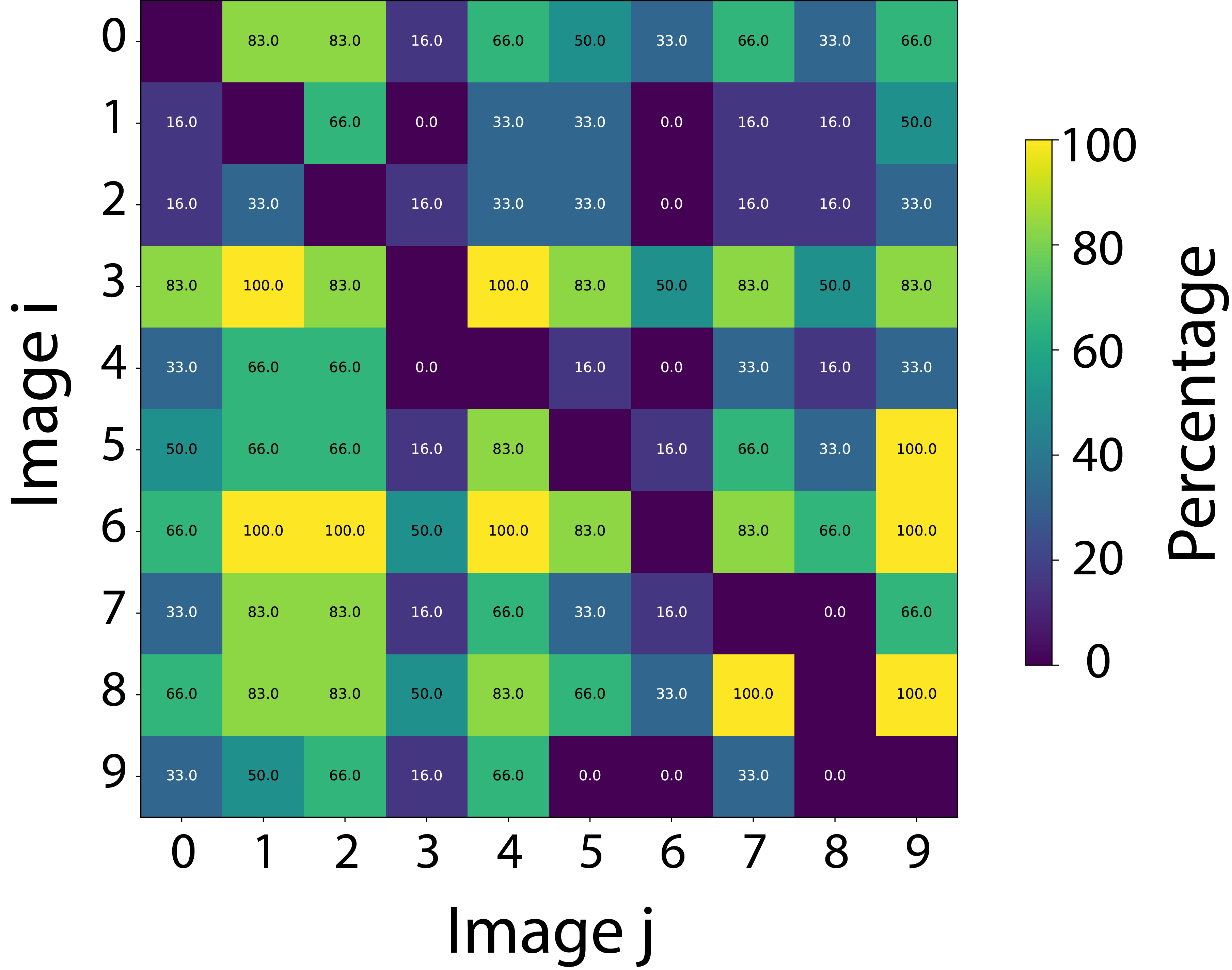}
\caption{\textbf{The data showed between-subject consistency}.
Each cell (i, j) represents the percentage of times when image i was rated funnier than image j. Participants tended to agree on which image was funnier, showing images 6 and 3 being rated funnier than others most times. 
}
\label{fig:direct_compare_across}
\end{figure}

\begin{figure}
\centering
\includegraphics[height=0.5\linewidth]{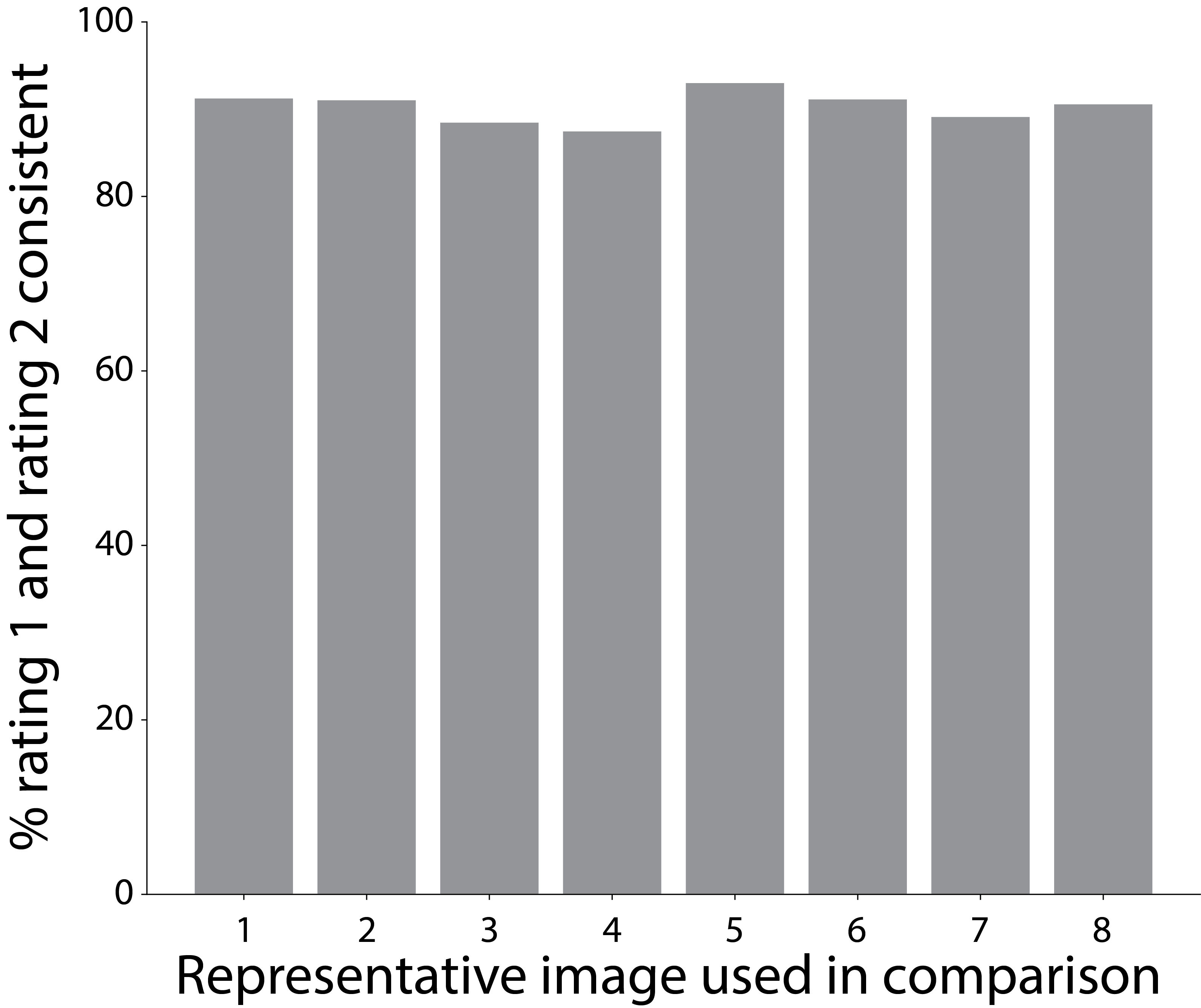}
\caption{\textbf{Participants showed self-consistency in the Comparison task.}
The x-axis shows the representative images, while the y-axis shows the percentage of instances where a user's second rating matched their first for comparison containing the particular comparison image. A 100\% match is perfect self-consistency.}
\label{fig:direct_comppare_within}
\end{figure}

\noindent \textbf{Reliability between participants}.
We assessed between-participant agreement for the comparison task through:
\begin{itemize}
\item {Anchor Image Comparisons:} There was high reliability when comparing results against anchor images (\textbf{Fig.\ref{fig:direct_comppare_within}})
\item {Comprehensive Pilot:} There was strong agreement in all possible pairwise comparisons of 10 random images (different from anchor images) ($\binom{10}{2}$ pairs, \textbf{Fig.\ref{fig:direct_compare_across}})
\end{itemize}

\subsubsection*{Final Dataset}
Each image in our 3,542-image set received 6-8 ratings (Binary or Range tasks) or 4-6 comparisons. Of these, 1,771 images were labeled ``funny'' (mean binary rating $\geq$ 0.5). To ensure balance, we removed additional non-funny images with the highest rating variance. For training/validation/test splits, we ensured each set contained both original and modified versions of any given image pair to avoid biases. \textbf{Table \ref{tab:dataset_summary}} summarizes the final dataset composition.

We conducted two separate evaluations: one on the entire test set (testAllSet) and the other on the subset of the test set consisting of only the original/modified pairs of images in the test set (testOnlyPairs).

\begin{table*}
\centering
\small
\begin{tabularx}{\textwidth}{|Y|Z|Z|Z|Z|Z|Z|Z|Z|}
\hline
Dataset & Total & Pairs & \centering{Funny images} & & & \centering{Not Funny} & & \\
\hline
 & & & N & Binary & Range & N & Binary & Range \\
\hline
Training& $2,136$ & 698 & 1,068 & $0.79\pm0.18$ & $5.75\pm1.42$ & 1,068 & $0.12\pm0.16$ & $3.62\pm1.75$\\
\hline
Valid.& $703$ & 273 & 351 & $0.78 \pm 0.19$ & $5.68 \pm 1.40$ & 351 & $0.12 \pm 0.17$ & $3.65 \pm 1.65$\\
\hline
testAll & $706$ & 300 & 352 & $0.77 \pm 0.19$ & $5.60 \pm 1.37$ & 352 & $0.14 \pm 0.17$ & $3.39 \pm 1.68$\\
\hline
testOnlyPairs & $600$ & 300 & 300 & $0.77\pm 0.19$ & $5.62 \pm 1.36$ & 300 & $0.13\pm 0.16$ & $3.30 \pm 1.68$\\
\hline
\textbf{Total} & 3,542 & 1,271 & 1,771 & $0.79 \pm 0.20$ & $5.70 \pm 1.40$ & 1,771 & $0.13 \pm 0.17$ & $3.58 \pm 1.72$\\
\hline
\end{tabularx}
\caption{Dataset summary. Images were labeled as humorous if the mean participant rating was $\geq$ 0.5. The dataset was balanced by removing non-funny images with the highest rating standard deviations. Splits (training, validation, test) contain both original and modified versions of each image to avoid bias. The table presents two test evaluations: testAllSet (entire test set) and testOnlyPairs (subset with original/modified pairs). Note: The Total excludes testOnlyPairs as it's a subset of testAllSet.}
\label{tab:dataset_summary}
\end{table*}

\section{Experiments}
\label{sec:expts}

\textbf{\large Models}

We evaluated state-of-the-art visual architectures, including vision-only and vision-language models, using both pretrained and non-pretrained settings. 
We evaluated the following models:
DINOv2 large \citep{oquab2023dinov2}, ViT huge \citep{dehghani2023scaling}, Swin2 large \citep{liu2022swin}, ConvNeXt large \citep{liu2022convnet}, ViTG-14 \citep{sun2023evaclip, li2023blip2}, ResNet152 \citep{he2015deep}, LLaVA (in zero-shot and fine-tuned configurations) \citep{liu2023improvedllava, liu2023llava}, BLIP (fine-tuned) \citep{li2022blip}, GPT-4o (gpt-4o-2024-05-13) \citep{openai2023gpt4}, and Gemini 1.5-002 Flash \citep{geminiteam2024gemini}.
Pretrained vision models were initialized with weights from either ImageNet \cite{russakovsky2015imagenet} or LAION-2B \cite{schuhmann2022laion5b}. BLIP and LLaVA,  use a combination of various datasets for training. GPT-4o and Gemini-Flash do not open-source their training data.

For vision-only models in the Binary and Range tasks, we added a final fully connected layer for predictions (funny/not funny and 1--10, respectively). For the Comparison task, we extracted features before the classification layer, computed the difference between the two images' features, and passed the difference to a classifier for prediction. 

For vision-language models, we framed our tasks as VQA problems. For the Binary task, the prompts were ``Is the image funny?'' and ``Is the image not funny?'', averaging results as performance was comparable between the two. For the Range task, we asked ``What is the degree of funniness of this image from 1 to 10?'' For the Comparison task in zero-shot setting, we prompted: ``Given these two images, answer which is funnier with 'first' or 'second' and explain why succinctly''. We confirmed that removing the word "succinctly" did not significantly alter performance.

We also used the words participants used to describe funny images (Comparison task). We selected words common in at least 30\% of the responses for each image. For fine-tuning, we modified the prompt by using a prefix: ``The prominent features of this image are: {common words}''.

Each experiment was conducted 5 times for all models except for GPT-4o and Gemini-Flash, which were run only. Further training details are shown in Appendix \cref{sec:train_detail}.

\begin{figure*}[!htbp]
\centering
\begin{subfigure}[t]{0.49\linewidth}
\includegraphics[height=0.6\linewidth]{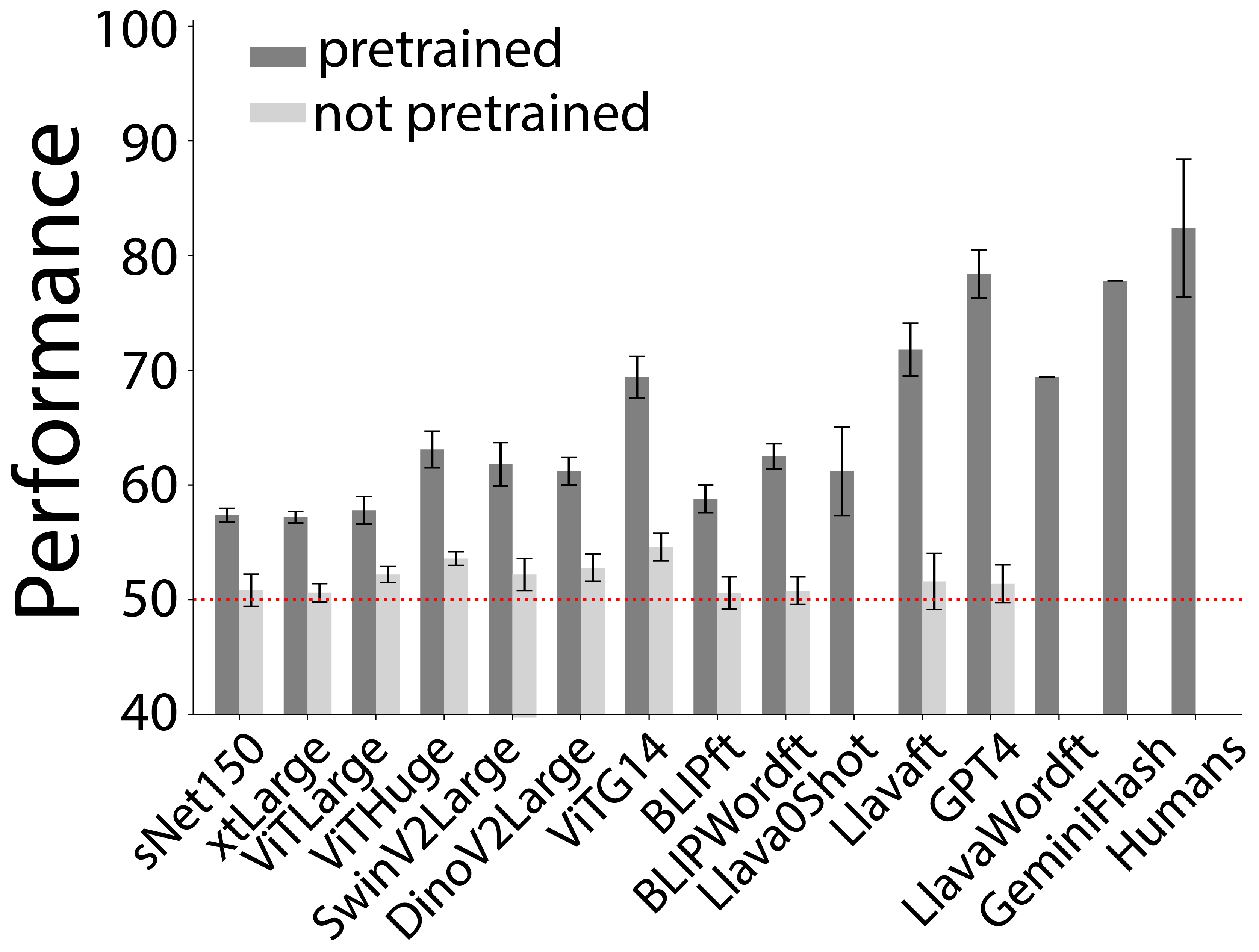}
\end{subfigure}
\hspace{0.2em} 
\begin{subfigure}[t]{0.49\linewidth}
\includegraphics[height=0.6\linewidth]{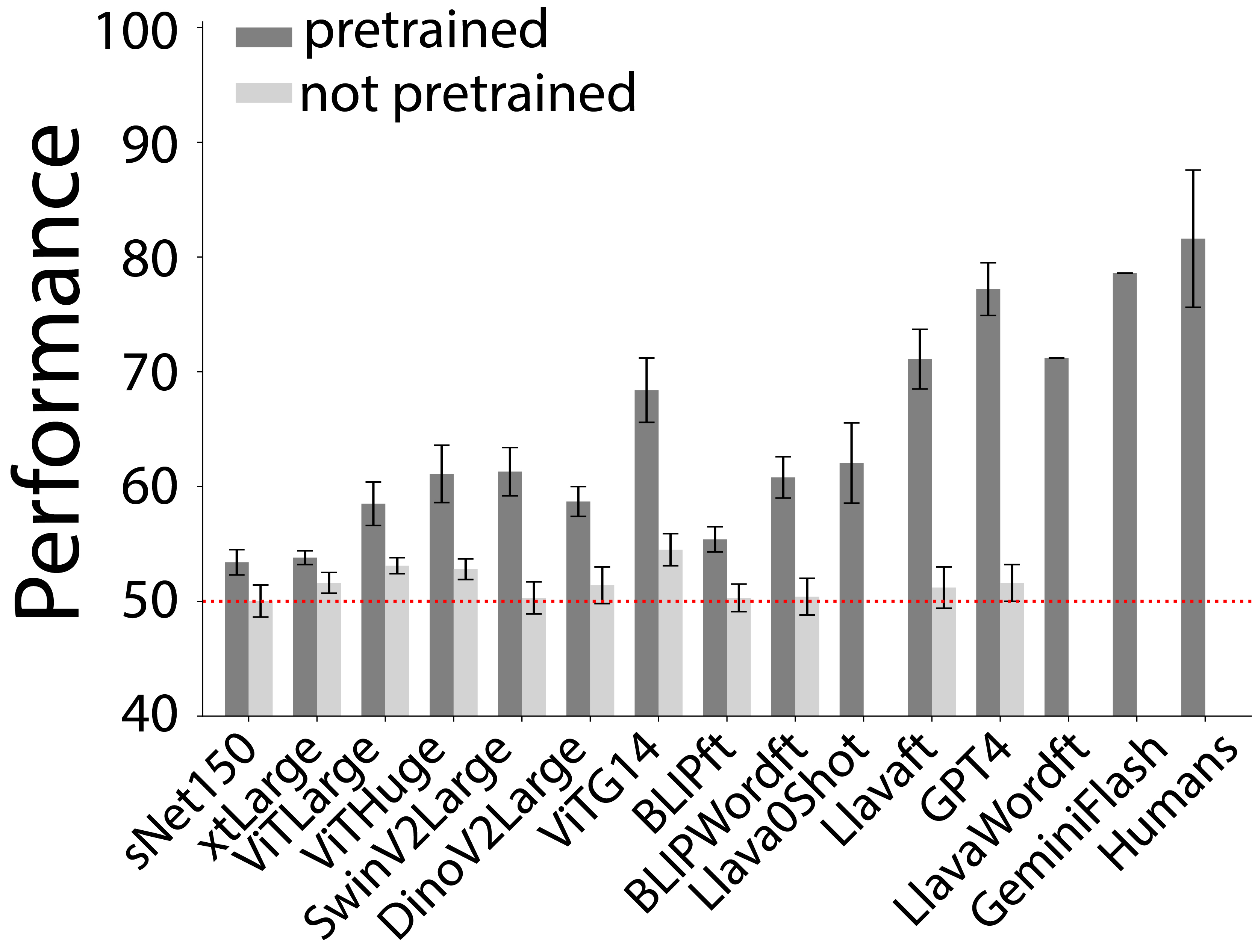}
\end{subfigure}
\caption{
\textbf{Binary task results for (a) \texttt{testAllSet} and (b) \texttt{testOnlyPairs}}. Dark grey bars represent pretrained models; light grey bars represent non-pretrained models. The dotted line indicates chance performance. The last column shows human performance. Error bars represent standard deviation. ``ft'' denotes fine tuning. ``Wordft'' denotes fine tuning and usage of word descriptors (see text for details).
}
\label{fig:binary_all_results}
\end{figure*}

\subsection*{Tasks}
\label{sec: tasks_descrip}

\noindent \textbf{Binary Task.}
We measured accuracy, comparing the predicted label (``funny'' or ``not funny'') against the average human-based label (threshold of 0.5). Chance was 50\%. To assess control for editing artifacts and model learning beyond HumorDB, we also evaluated model performance on an external set of 650 non-humorous, non-edited images (MSCOCO\cite{lin2015microsoftcococommonobjects}: 250, Places365\cite{Places365data7968387}: 400).

\noindent \textbf{Range Task.}
We computed the root mean square error (RMSE) between the predicted rating (1--10) and the mean human rating. Chance levels were estimated under two null hypotheses: (i) sampling random labels from the empirical distribution of funniness ratings in the dataset, (ii) sampling uniform random integers in 1--10 (for zero-shot models).

\noindent \textbf{Comparison Task.}
Accuracy was computed as the fraction of pairs correctly identified as ``A is funnier'' or ``B is funnier,'' matching the majority human vote. Chance was 50\%.

\noindent \textbf{Testing VLM Explanations}
We conducted a human evaluation of VLM explanations Participants rated explanations on a 1--5 scale based on: (1) how well they identified \emph{why} an image is funny, and (2) factual accuracy of scene. We also used an automated mechanism by checking the presence of word synonynms from human annotations using: (1) All Words Score - considering all annotator words, and (2) Common Words Score - focusing on words appearing in at least 30\% of human responses.
(see Appendix \cref{sec:score_explain}).

\noindent \textbf{Attention Maps}
We analyzed attention maps in vision transformers using attention rollout \cite{abnar2020quantifying}. We produced global attention maps over image tokens for test images.
For paired images, we approximated the humorous region using pixel-level differences between original and modified versions. We evaluated attention map quality using recall, strict box containment, and outside box ratio (Appendix~\cref{sec:attn_map_eval}).

\noindent \textbf{Mechanistic Interpretability}
We adapted logit attribution to evaluate how each transformer layer contributed to humor detection. We treated each ViT block as a layer and measured classification accuracy using each layer's predictions (Appendix \cref{sec:logit_attr}).
\section{Results}
\label{sec:results}
We evaluated models on three tasks: (1) Binary Classification, (2) Range Rating, and (3) Comparison, using both the full test set (\texttt{testAllSet}) and a subset containing only paired images (\texttt{testOnlyPairs}, e.g., \textbf{Fig.~\ref{fig:image_pair}}).

\noindent \textbf{Binary Task}.
\label{subsec:binary}
As expected, pretrained models (dark bars) consistently outperformed non-pretrained models (light bars) both for
\texttt{testAllSet} (\textbf{\Cref{fig:binary_all_results}a}) and \texttt{testOnlyPairs}  (\textbf{\Cref{fig:binary_all_results}b}). The latter generally performed near chance, demonstrating the importance of pretraining priors. Performance dropped when evaluating only on paired images, emphasizing that such contrastive examples pose a more stringent test for visual tasks. Large vision-language models (e.g., Gemini-Flash, fine-tuned LLaVA) outperformed vision-only models, while zero-shot large multimodal LLMs also performed surprisingly well despite no \emph{specific} humor training (though these models have likely been exposed to humorous content). Higher-capacity models maintained relatively stable performance from \texttt{testAllSet} to \texttt{testOnlyPairs}, whereas smaller models showed a bigger drop. Finally, vision-language models (LLaVA, BLIP \cite{li2022blip}) trained with supporting words for a portion of the images further enhanced performance, indicating the value of human-guided training.

On the control set of not funny images from external datasets, 
models classified images as ``not funny'': ViT-G14 (81\%), LLaVA FT (80\%), SwinV2 (76\%), ViTHuge (71\%), DinoV2 (62\%). Thus, models are learning beyond superficial editing cues and showing learning beyond HumorDB.

\noindent \textbf{Range Task}.
The strongest models achieved RMSE values around 1.7-2.0, significantly better than random baselines \textbf{Table \ref{tab:model_range_performance}}.
 Large language models performed reasonably well even in zero-shot settings, suggesting an inherent numeric humor scoring ability. Fine-tuned LLaVA variants (especially those trained with human keywords) performed best, verifying the usefulness of targeted training.

\begin{figure}[!htbp]
\centering
\includegraphics[width=.8\linewidth]{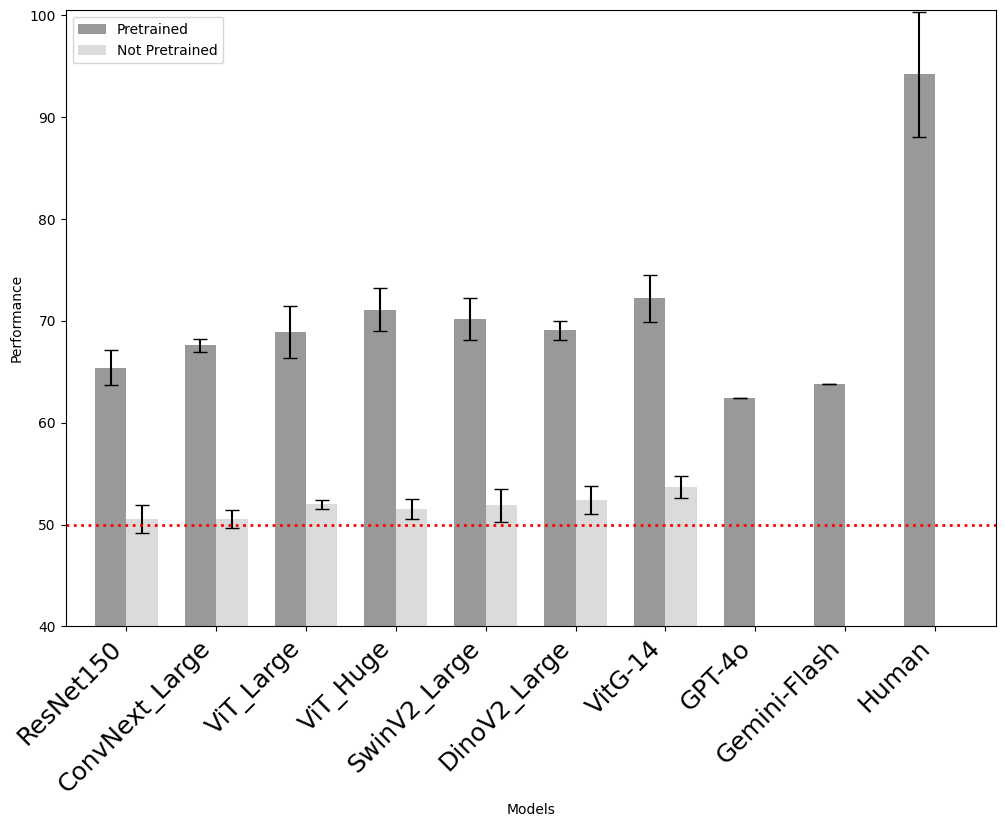}
\caption{\textbf{Comparison task performance} on \texttt{testAllSet}. Pretrained models still trail far behind humans.}
\label{fig:compare_all_results}
\end{figure}

\noindent \textbf{Comparison Task}.
Choosing which of two images is funnier proved more challenging (\textbf{Fig. \ref{fig:compare_all_results}}). Although performance exceeded chance levels (50\%), it was notably lower than on the Binary task, highlighting the additional complexity of making nuanced judgments between similar images. Interestingly, models trained directly on comparison data outperformed larger zero-shot LLMs in this task.

\noindent \textbf{Performance Across Image Categories}.
Performance varied markedly by image type (\ref{tab:binary_types}). All models performed near chance on sketches, likely due to their abstract nature. Performance on photoshopped and AI-generated images was strongest, perhaps because these images share characteristics with the models' training data.

\begin{figure}[!htbp]
\centering
\includegraphics[height=0.2\textheight]{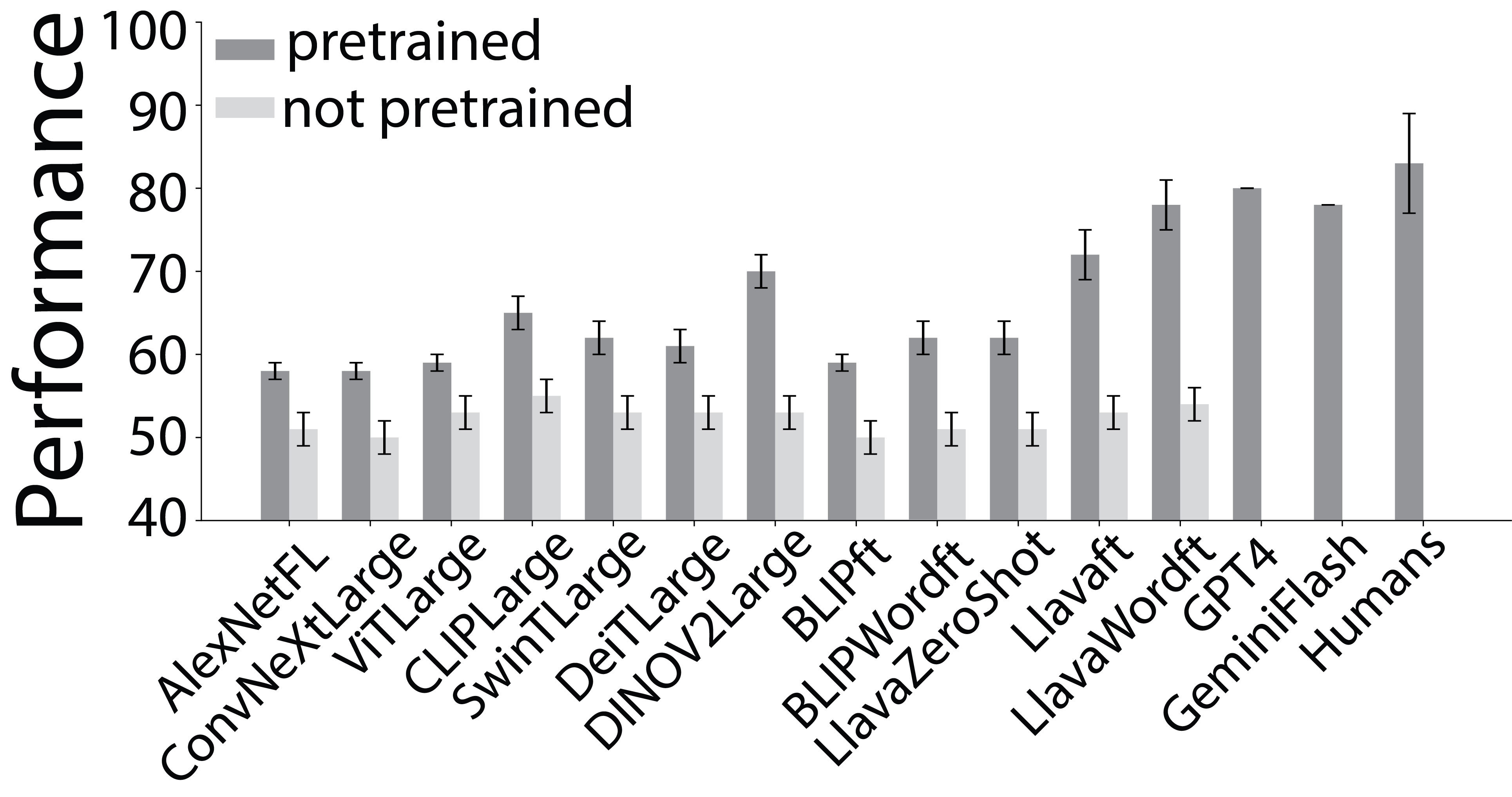}
\caption{\textbf{Binary task results evaluated only with original images in the testAllSet} (i.e., excluding testing on modified images). Format and conventions as in \textbf{Fig.~\ref{fig:binary_all_results}}. Several models improve their performance on the original images, indicating that the modified images serve as an important control.
GPT-4o shows a particularly surprising increase in performance suggesting that it probably has seen the original images from the internet in its training set.
}
\label{fig:result_org_only}
\end{figure}

\noindent \textbf{Importance of Minimally Contrastive Pairs}.
When evaluated only on original images (excluding modified pairs), several models showed significant performance increases (\textbf{Fig. \ref{fig:result_org_only}}). GPT-4o exhibited a surprising 10\% improvement, reinforcing the necessity of contrastive pairs for rigorous evaluation and suggesting potential exposure to similar internet images during training.

\noindent \textbf{Explanation Quality}.
\noindent \textbf{Explanation Quality.} Our human evaluation of VLM explanations revealed strong performance, particularly from larger models. On a 1--5 scale, Gemini's explanations were rated highest for both identifying the source of humor ($4.1\pm0.6$) and for overall accuracy ($4.24\pm0.47$). GPT-4o also performed well (Why Funny: $3.9\pm0.7$, Accuracy: $4.12\pm0.63$), followed by LLaVA (Why Funny: $3.5\pm0.9$, Accuracy: $3.57\pm0.95$). These results suggest a robust, though not perfect, ability to articulate visual humor beyond simple keyword matching. Automated scores are detailed in Appendix \ref{sec:score_explain}.

\begin{figure*}[h]
\vspace{5pt}
\centering
\begin{subfigure}{0.24\textwidth}
  \includegraphics[width=0.9\linewidth]{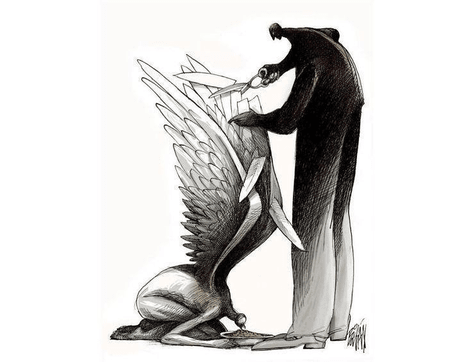}
  \caption{LLM response: \emph{no. this image appears to depict a somber or serious theme. the artistic representation of a person cutting another individual's wings suggests a metaphor for limiting someone's freedom or potential}. Range rating: 1}
\end{subfigure}
\hfill
\begin{subfigure}{0.24\textwidth}
  \includegraphics[width=0.9\linewidth]{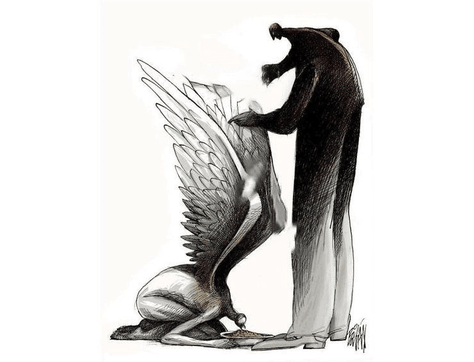}
  \caption{LLM response: \emph{no, the image is not funny. it appears to be a serious, thought-provoking illustration that likely carries a significant meaning or message. the imagery of a person putting a hood over a winged figure}. Range rating: 1}
\end{subfigure}
\hfill
\begin{subfigure}{0.24\textwidth}
  \includegraphics[width=0.9\linewidth]{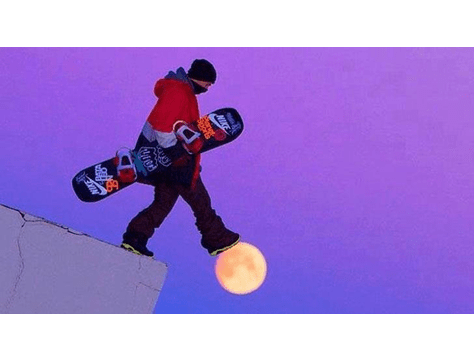}
  \caption{LLM response: \emph{yes. the image is funny because it shows a snowboarder appearing to step on the moon, playing with perspective to create a humorous and surreal visual effect}. Range rating: 7}
\end{subfigure}
\hfill
\begin{subfigure}{0.24\textwidth}
  \includegraphics[width=0.9\linewidth]{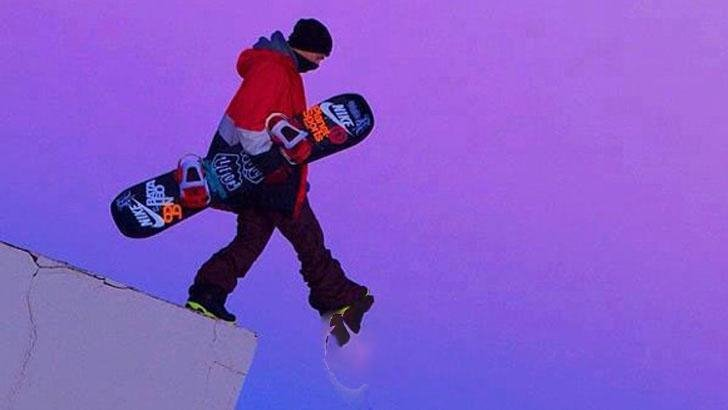}
  \caption{LLM response: \emph{no. the image depicts someone who appears to be about to step off a ledge while holding a snowboard, which could be interpreted as potentially dangerous rather than funny}. Range rating: 1}
\end{subfigure}

\vspace{2em} 
\begin{subfigure}{0.24\textwidth}
  \includegraphics[width=0.9\linewidth]{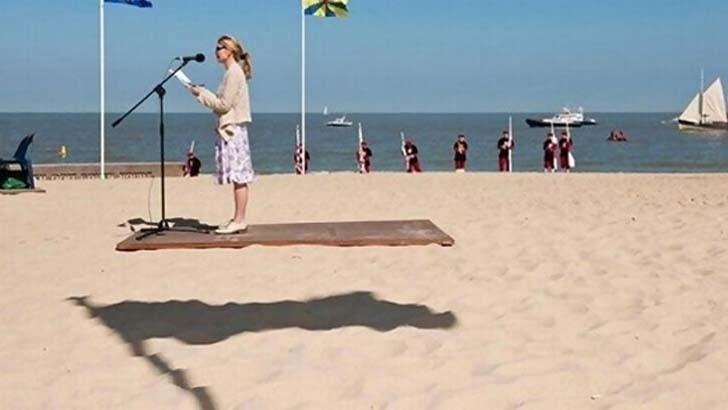}
  \caption{LLM response: \emph{yes, the image can be considered funny due to the optical illusion created by the shadow on the beach, which makes it appear as though someone is flying on a magic carpet}. Range rating: 9}
\end{subfigure}
\hfill
\begin{subfigure}{0.24\textwidth}
  \includegraphics[width=0.9\linewidth]{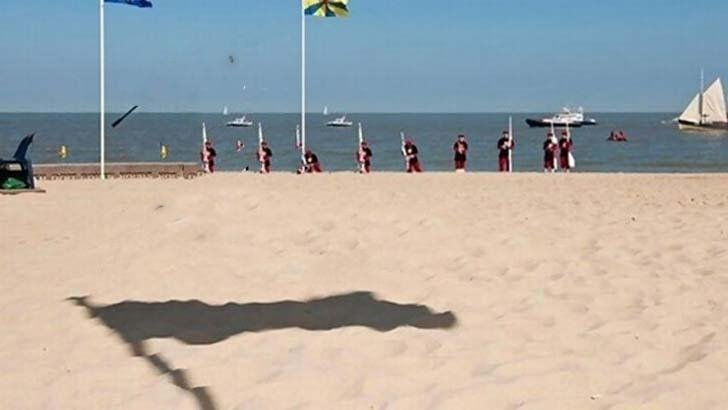}
  \caption{LLM response: \emph{yes. the image is funny due to the optical illusion caused by the shadow of the flagpole, which makes it appear as though there is a large figure casting a long shadow on the beach}. Range rating: 7}
\end{subfigure}
\hfill
\begin{subfigure}{0.24\textwidth}
  \includegraphics[width=0.9\linewidth]{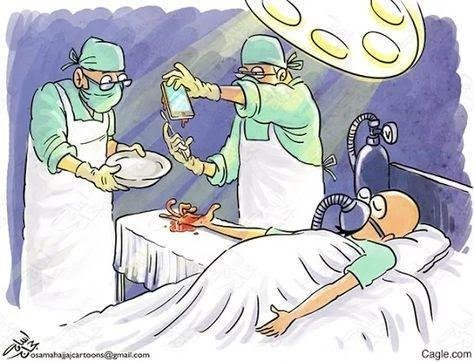}
  \caption{LLM response: \emph{yes, the image is funny.
Explanation: The cartoon depicts a surgical operation where the surgeons, instead of using typical medical equipment, appear to be treating the patient as if they are preparing food.} Range rating: 8}
\end{subfigure}
\hfill
\begin{subfigure}{0.24\textwidth}
  \includegraphics[width=0.9\linewidth]{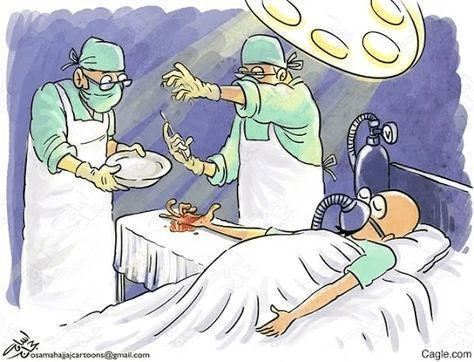}
  \caption{LLM response: \emph{Yes. The humor comes from a surgeon theatrically throwing the patient's internal organs like a basketball while another surgeon waits to catch them with a plate}. Range rating: 8}
\end{subfigure}
\caption{\textbf{Example GPT-4 binary classification, range ratings, and explanations.} The human binary and range ratings are as follows for the images: (a) \emph{66\%, 4.20}, (b) \emph{0\%, 5.6}, (c) \emph{71\%, 6.70}, (d) \emph{0\%, 3.60}, (e) \emph{60\%, 2.80}, (f) \emph{0\%, 5.57}, (g) \emph{83\%, 6.75}, (h) \emph{14\%, 5.70}}
\label{fig:gpt_4_images}
\end{figure*}

\noindent \textbf{Qualitative Analysis}.
\textbf{Fig. \ref{fig:gpt_4_images}} shows examples of GPT-4o classification and explanations. The model sometimes labeled both paired images as funny, failing to grasp distinguishing elements (e.g., \textbf{Fig. \ref{fig:gpt_4_images}g-h}). In some cases, the model provided eloquent explanations aligned with human reasoning while in others, it missed subtle but critical elements (e.g., \textbf{Fig. \ref{fig:gpt_4_images}c-d)}).

\noindent \textbf{Attention Maps and Mechanistic Interpretability}.
SOTA Vision Transformers (ViTs) show limited ability to identify humorous regions in images. Across ViT variants, attention maps achieved modest recall (0.17--0.26) with the humorous regions, with ViTG-14 performing slightly better (0.27) and DINOv2-Large notably worse (0.006). Models generally struggled with precise attention containment. ViTG-14 led with 42\% of its attention staying within the humor bounding box, followed by ViT-Huge (38\%) and ViT-Large (31\%). While DINOv2-Large showed higher containment (55\%), this was misleading given its near-zero recall. All models exhibited similar ratios of attention spillage outside the target box (0.30--0.36), indicating consistent difficulty in focusing specifically on humorous elements.

\begin{figure}
    \centering
    \includegraphics[height=0.6\linewidth]{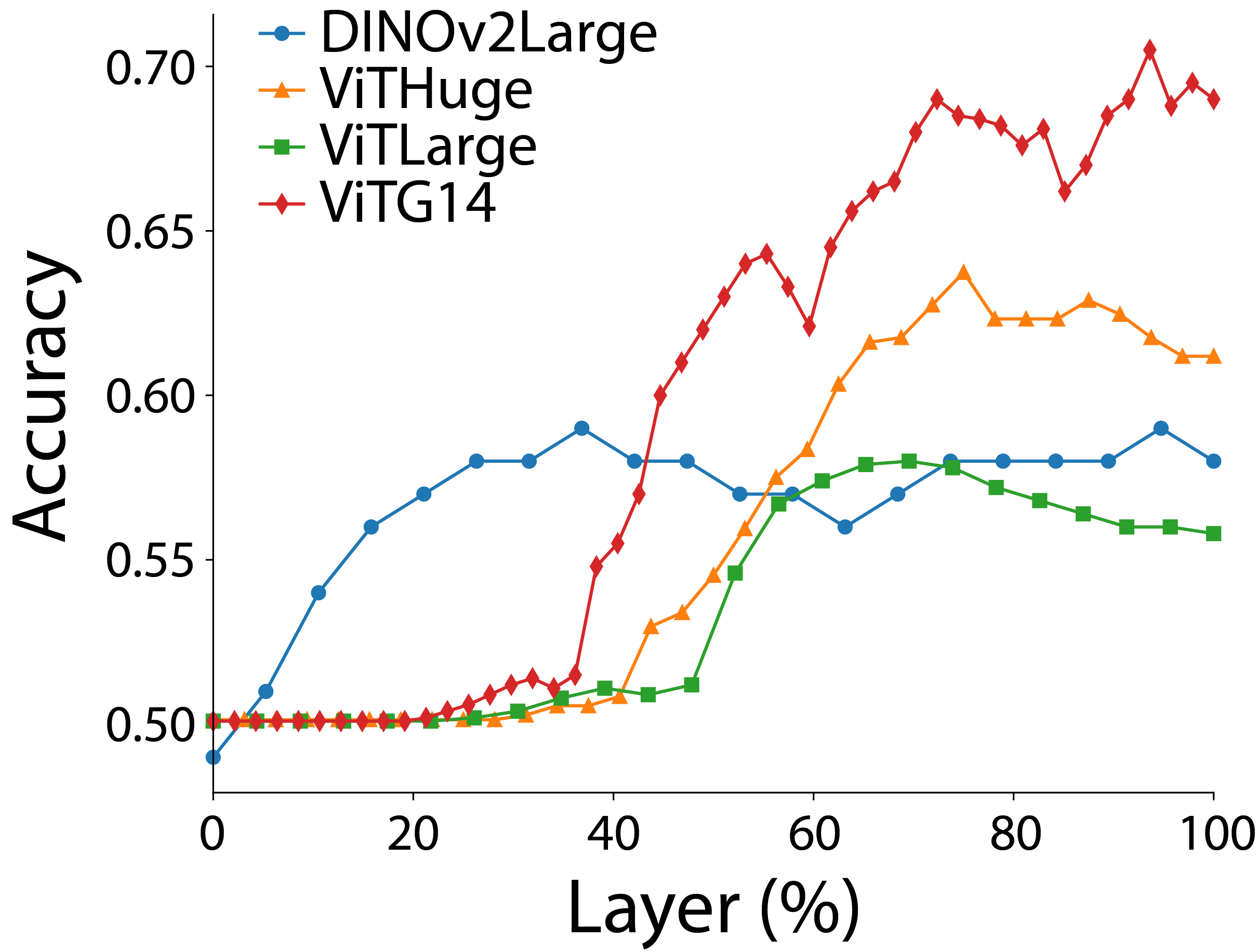}
    \caption{\textbf{Layerwise binary classification results}}
    \label{fig:enter-label}
\end{figure}

The logit attribution analysis revealed distinct patterns in how different architectures process humor across their network depths (\textbf{Fig.~\ref{fig:enter-label}}). The ViT-G14, the largest model tested, showed the most dramatic improvement in humor detection capability through its layers, reaching peak accuracy of ~70\% in its final layers. This suggests that humor understanding in this model emerges gradually through successive transformations of the visual features.
In contrast, DINOv2-Large exhibited early learning with rapid improvement in the first 20\% of layers but plateaued around 58\% accuracy, maintaining relatively consistent performance through deeper layers. ViT-Huge and ViT-Large showed similar patterns, with minimal learning in early layers (0-40\%) followed by steady improvement. However, their performance peaked lower than ViT-G14, at around 62\% and 57\% respectively. Both models showed slight performance degradation in their final layers, suggesting possible overfitting or feature collapse at extreme depths.

\begin{table}[!ht]
\centering
\small
\setlength{\tabcolsep}{3.5pt}
\renewcommand{\arraystretch}{0.9}
\caption{Attention Evaluation: Comparison of Selected Metrics (Recall, Strict Box Containment, Outside Box Ratio)}
\label{tab:attn_comparison}
\begin{tabular}{lcccccc}
\toprule
& \multicolumn{2}{c}{\textbf{Recall}} & \multicolumn{2}{c}{\textbf{SBC}} & \multicolumn{2}{c}{\textbf{OBR}} \\
\cmidrule(lr){2-3} \cmidrule(lr){4-5} \cmidrule(lr){6-7}
& Mean & Std & Mean & Std & Mean & Std \\
\midrule
ViT-L & 0.240 & 0.268 & 0.308 & 0.465 & 0.350 & 0.401 \\
ViT-H & 0.170 & 0.250 & 0.378 & 0.487 & 0.355 & 0.420 \\
DINOv2 & 0.006 & 0.013 & 0.552 & 0.502 & 0.324 & 0.440 \\
ViTG-14 & 0.265 & 0.218 & 0.425 & 0.497 & 0.348 & 0.381 \\
\bottomrule
\end{tabular}
\end{table}

\begin{table*}
\centering
\begin{tabularx}{\textwidth}{|X|X|X|}
\hline
Model Name & testAllSet RMSE $\pm SD$ & testOnlyPairs RMSE $\pm SD$\\
\hline
dinov2 large & $1.98 \pm 0.08$ & $1.96 \pm 0.09$\\
\hline
vit huge & $1.88 \pm 0.09$ & $1.92 \pm 0.07$\\
\hline
swin2 large & $1.96 \pm 0.12$ & $1.98 \pm 0.11$\\
\hline
convnext large & $2.10 \pm 0.06$ & $2.15 \pm 0.07$\\
\hline
vitg 14 & $1.76 \pm 0.08$ & $1.80 \pm 0.06$\\
\hline
resnet152 & $2.11 \pm 0.09$ & $2.09 \pm 0.11$\\
\hline
LLaVA(Zero-Shot) & $2.95 \pm 0.33$ & $2.98 \pm 0.34$\\
\hline
LLaVA(fine-tuned) & $1.70 \pm 0.22$ & $1.69 \pm 0.22$\\
\hline
LLaVA(words fine-tuned) & $\mathbf{1.68 \pm 0.28}$ & $\mathbf{1.66 \pm 0.31}$\\
\hline
BLIP (fine-tuned) & $1.94 \pm 0.05$ & $1.96 \pm 0.05$\\
\hline
BLIP (words fine-tuned) & $1.92 \pm 0.06$ & $1.95 \pm 0.04$\\
\hline
GPT-4o(Zero-Shot) & $2.61$ & $2.63$\\
\hline
Gemini-Flash & $2.06$ & $2.11$\\
\hline
Humans & $2.72 \pm 0.88$ & $2.71 \pm 0.86$\\
\hline
Chance(from distribution) & $2.69 \pm 0.05$ & $2.50 \pm 0.06$\\
\hline
Chance(for zero-shot) & $3.58 \pm 0.07$ & $3.33 \pm 0.07$\\
\hline
\end{tabularx}
\caption{Range Task Results on testAllSet and testOnlyPairs. Bold indicates the best model result. The chance(from distribution) scores are calculated by random sampling from the distribution of ratings in the dataset after rounding the mean range ratings to the nearest integer. The chance(zero-shot) scores are calculated after randomly sampling an integer from 1 to 10.}
\label{tab:model_range_performance}
\end{table*}

\section*{Discussion}

\noindent \textbf{Performance Gaps}. All models performed above chance levels in the Binary task but were below human accuracy despite fine-tuning. This gap was more pronounced in the Comparison task, underscoring the nuanced visual reasoning required for humor. Interestingly, in the Range task, some models achieved performance comparable to humans.

\noindent \textbf{Value of Contrastive Pairs}. The minimally contrastive pairs proved crucial for exposing model limitations. Performance consistently dropped on such pairs, revealing that many models rely on superficial cues rather than identifying precise humor-inducing elements. Notably, when evaluated only on original images, several models (particularly GPT-4o) showed significant performance increases, suggesting potential familiarity with internet-sourced funny images.

\noindent \textbf{Image Type Variation}. Performance varied across categories: models handled photoshopped and AI-generated content better but struggled with sketches, highlighting how abstraction challenges humor recognition.

\noindent \textbf{Large Model Advantages}. Large vision-language models achieved higher accuracy than vision-only models, suggesting broad pretraining aids in detecting humor-related cues. Zero-shot large multimodal LLMs (GPT-4o, Gemini-Flash) displayed promising capabilities, though they still made systematic errors on challenging examples and showed inconsistencies in explanations.

\noindent \textbf{Understanding vs. Classification}. While models often classified images correctly, deeper analyses revealed limitations. Attention maps showed that even when models were correct, they rarely focused precisely on the humorous regions. Model explanations sometimes aligned with human annotations but often hallucinated reasons or highlighted irrelevant details, particularly for contrastive pairs. This disconnect between classification and ``understanding'' highlights the ongoing challenge in grounding visual-linguistic representations in genuinely human-like reasoning.

\noindent \textbf{Limitations}. Our dataset is comparable to established benchmarks in terms of \#images per class, but it is conceivable, even likely, that more data would improve performance. Substantially enlarging \textbf{HumorDB} would introduce logistical hurdles, such as additional image modifications, curation, and participant annotations. Moreover, humor is highly culture-dependent; thus, achieving a fully global perspective would necessitate broader sampling across numerous cultural contexts and humor styles. We expect that the observed gap in model performance—especially in the Binary and Comparison tasks—is driven less by data quantity than by the inherent challenges in abstract visual reasoning required for humor. Evidence of this limitation appears in \textbf{Fig.~\ref{fig:gpt_4_images}} and Appendix \cref{sec:atten_maps}, where GPT-4o explanations and ViT-Huge attention maps fail to highlight critical incongruities (e.g., overlooking the cell phone in  \textbf{Fig.~\ref{fig:image_pair}}). Such omissions emphasize the difficulty of pinpointing subtle yet crucial cues that distinguish ``funny'' from ``not funny.''

Furthermore, while our participant pool was diverse (see Appendix~\ref{sec:participant_demo}), it was predominantly from the United States (54\%) and younger demographics (<40 years). As humor is culture-dependent, future work could improve upon the work by considering a stratified demographic group.

Using \textbf{HumorDB} as a zero-shot benchmark for large multimodal models could shed light on how these models generalize to complex, abstract tasks like humor. Our dataset construction methodology including \emph{minimally-contrastive} pairs may also benefit other domains to isolate subtle semantic changes. As the field moves from labeling individual objects to more nuanced scene interpretation, accounting for human variation becomes crucial (e.g., one viewer’s ``funny'' might be another’s ``not funny''). Although \textbf{HumorDB} focuses on majority votes and average ratings, future expansions could gather per-user data and probe the subjective core of humor. Such refinements would deepen our grasp of subjective labeling, cultural dependencies, and the abstract reasoning that still eludes current AI systems (see also Appendix\cref{sec:future_dirs}).

\newpage
\section{Acknowledgements}
The authors would like to thank Morgan Talbot and Spandan Madan for their help with the GPU cluster and running experiments, the members of Kreiman Lab for the In-Lab study and weekly discussions.

This research was supported by the Illinois Computes project at the University of Illinois Urbana-Champaign (JV), NIH Grant R01EY026025 (GK), and the Center for Brains, Minds and Machines at Harvard/MIT (GK). 
{
    \small
    \bibliographystyle{ieeenat_fullname}
    \bibliography{main}
}

\clearpage
\setcounter{page}{1}
\maketitlesupplementary

\section{Appendix}

\subsection{Training details}
\label{sec:train_detail}

Models were trained using the Adam optimization algorithm with weight decay. We conducted a hyperparameter grid search across learning rates in the set {0.01, 0.001, 0.0001, 0.00001}, batch sizes in the set {4, 8, 16}, and weight decay parameters in the set {0.1, 0.01, 0.001}. Model training proceeded for a fixed number of 10 epochs, with periodic checkpoints. For the final evaluation on the unseen test set, we used the model iteration exhibiting optimal performance on the validation set.
We used cross-entropy loss for the Binary classification and Comparison tasks, while mean square loss was used for the Regression task. For all architectures except GPT-4o and Gemini-Flash, we fine-tuned the models. For LLaVA, we performed LoRA fine-tuning instead of full fine-tuning.
Due to the uneven distribution of range ratings, we employed a sampling strategy that grouped images into bins according to their funniness ratings. This allowed us to randomly select a balanced number of images from each bin for every training epoch, ensuring a uniform distribution of sample images across all ratings in the training set. We applied the same sampling strategy for the Comparison task, which also contained a slightly uneven distribution.

Models were trained using the Adam optimization algorithm with weight decay. A hyperparameter grid search was conducted across learning rates in the set \(\{0.01, 0.001, 0.0001, 0.00001\}\), batch sizes in the set \(\{4, 8, 16\}\), and weight decay parameters in the set \(\{0.1, 0.01, 0.001\}\). Model training proceeded for a fixed number of 10 epochs, with periodic checkpoints. For the final evaluation on the unseen test set, we used the model iteration exhibiting optimal performance on the validation cohort. We used cross-entropy loss for the  Binary classification and Comparison tasks while mean square loss was used for the Regression task. For all architectures except GPT-4o and Gemini-Flash, we fine-tuned the models(Pretraining details are mentioned in Section 4.1). For LlaVA we did lora fine-tuning instead of full fine-tuning. To ensure statistical robustness, each experiment was conducted 5 times for all the experiments except for GPT-4o and Gemini-Flash which were run only one time.

Most of the experiments were run on 4 Nvidia GeForce RTX 2080 Ti GPUs which were part of an internal cluster. However, for training LLaVA and some models for the  Comparison task, we used an Nvidia A100 GPU. 

\subsection{External assets used}
\label{sec:external_asset}

We utilized the following assets:
The LLaVA repository (Apache-2.0 license) \citep{liu2023llava, liu2023improvedllava}, PyTorch \citep{pytorch}, huggingface transformers (Apache-2.0 license) \citep{wolf2020huggingfaces}, and huggingface accelerate (Apache-2.0 license) \citep{accelerate}.

Additionally, for the images collected from the internet we provide reference links in the repository.

\subsection{Attention maps}
\label{sec:atten_maps}

\begin{figure*}[h]
    \centering
    \includegraphics[height=0.25\textheight]{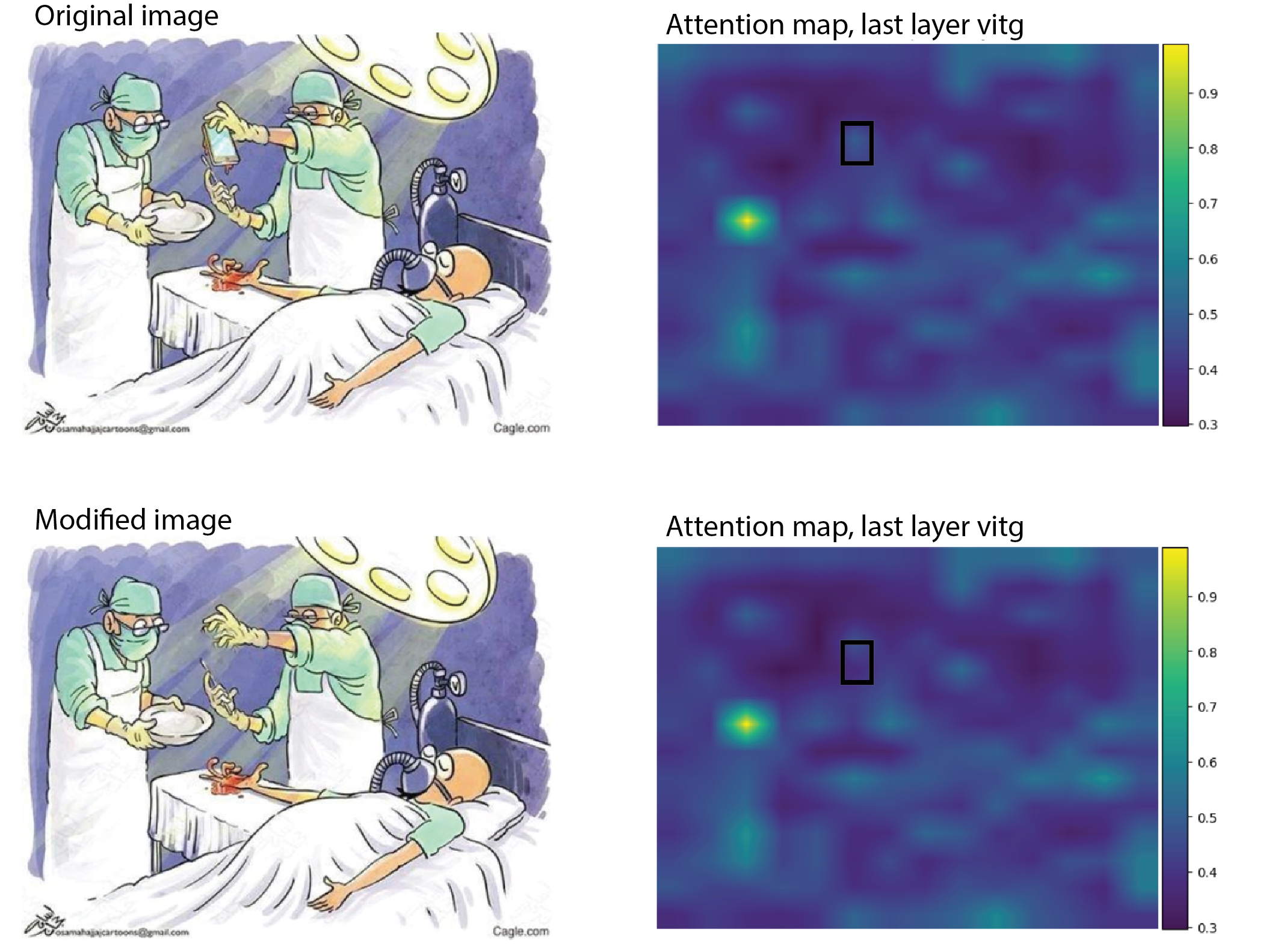}
    \caption{\textbf{Attention maps fail to capture elements critical to discern humor}. Attention maps based on the last layer of the vit huge model for the example images from \textbf{Fig.~\ref{fig:image_pair}}. 
   The black rectangle in the attention maps indicates the location of the phone. The maximum attention activation highlights the plate, which does not help distinguish between the original and modified images. Indeed, the model classified both images as funny.
   }
    \label{fig:image_1_attention}
\end{figure*}

We examined the attention maps using the attention rollout technique \cite{abnar2020quantifying} on the ViT-Huge model \cite{dehghani2023scaling}. This helped us understand whether the models focused on the actual humorous parts of images or other biases in the dataset. The attention maps may help to better understand how the models classify the images and identify potential shortcomings (\textbf{Fig.~\ref{fig:image_1_attention}}).

As an example, consider the case of \textbf{Fig.~\ref{fig:image_pair}}. The attention maps for the vit huge model are shown in \textbf{Fig.\ref{fig:image_1_attention}}. The model fails to pay attention to the most humorous part of the image (the phone, black rectangle), which is critical to assess whether the image is funny or not. Therefore the model is not able to correctly classify both images.

\subsection{Crowdsourcing details}
\label{sec:crowd_source_detail}

\begin{figure*}[h]
\vspace{-5pt}
\centering
\begin{subfigure}{0.3\textwidth}
  \includegraphics[width=\linewidth]{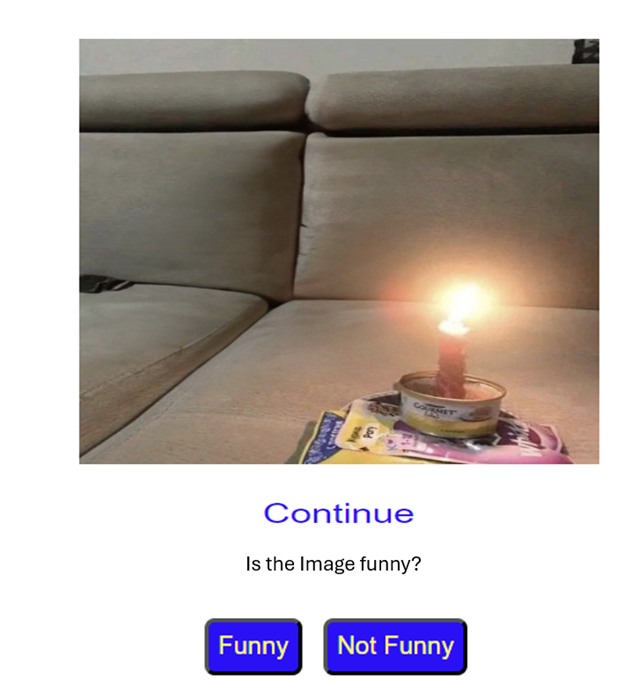}
  \caption{Binary Task Interface}
\end{subfigure}
\hfill
\begin{subfigure}{0.3\textwidth}
  \includegraphics[width=\linewidth]{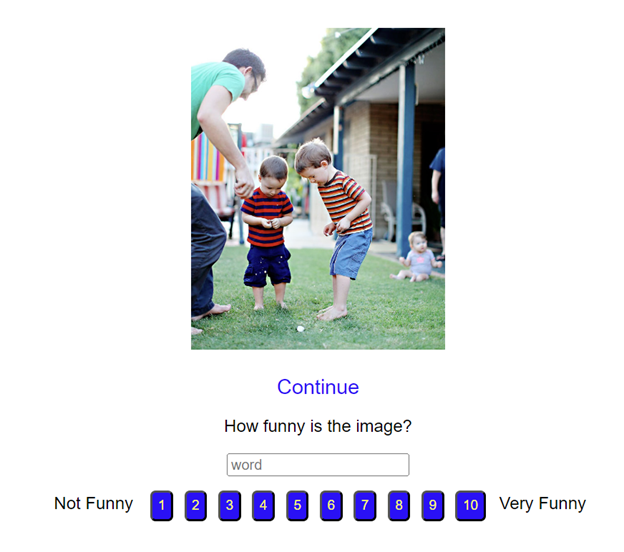}
  \caption{Range Task Interface}
\end{subfigure}
\hfill
\begin{subfigure}{0.3\textwidth}
  \includegraphics[width=\linewidth]{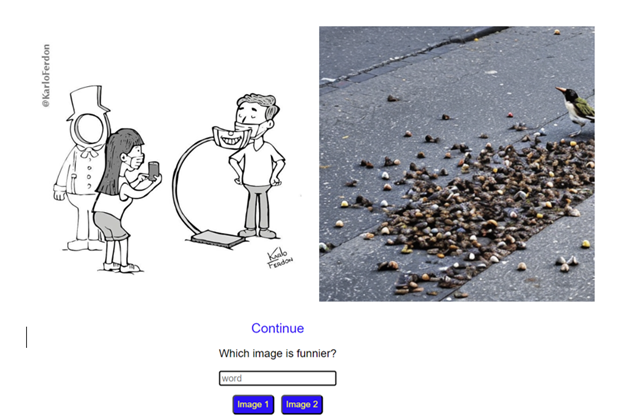}
  \caption{Compare Task Interface}
\end{subfigure}
\caption{\textbf{Crowd sourcing interfaces for the three tasks.}}
\label{fig:crowd_source}
\end{figure*}

There were 850 participants: 200 for binary task, 215 for the range task and 435 for the comparison task. The interfaces used by the participants for the three tasks are shown in \textbf{Fig.~\ref{fig:crowd_source}}. The generic instructions given for all tasks were:

\begin{itemize}
\item \textbf{Binary Task:} Please rate if the image is funny or not.
\item \textbf{Range Task:} Please rate the degree of funniness of the image on a scale from 1 (not funny) to 10 (very funny). 
\item \textbf{Comparison Task:} Please indicate which of the two images is funnier.
\item For funny images write a word that makes the image funny, for not funny images, write a word about the most prominent feature of the image.
\item The time required to rate all the images is approximately 9-11 minutes
\item Only click on the rating buttons once, and wait till the next image loads (maximum 1 second), a message will show you when the next image is being processed.
\item Please do not refresh the page. You will lose progress and will have to start again.
\item There are 100-120 images in this survey.
\item At the end of the survey, we will provide you a code, please store it and use it appropriately to get the reward.
\item Click the button below to begin.
\end{itemize}

Some participants were discarded due to reliability and the amount of outliers in their responses as detailed in Section3.2. All participants were required to submit all questions in the survey for the response to count.

\subsection{Multimodal models' answer explanations}
\label{sec:multi_explain}

\begin{figure*}[h]
\vspace{5pt}
\centering
\begin{subfigure}{0.24\textwidth}
  \includegraphics[width=\linewidth]{figures/gpt7.png}
  \caption{LLM response: yes. the humor lies in the fact that the patient's hand is being operated on with a smartphone. Range rating: 7}
\end{subfigure}
\hfill
\begin{subfigure}{0.24\textwidth}
  \includegraphics[width=\linewidth]{figures/gpt8.png}
  \caption{LLM response: yes. the patient's hand is detached from their body. Range rating: 8}
\end{subfigure}
\hfill
\begin{subfigure}{0.24\textwidth}
  \includegraphics[width=\linewidth]{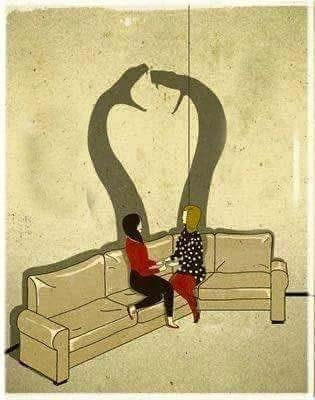}
  \caption{LLM response: yes. the shadow of the women looks like two snakes.  it is a dark and ironic humor. Range rating: 7}
\end{subfigure}
\hfill
\begin{subfigure}{0.24\textwidth}
  \includegraphics[width=\linewidth]{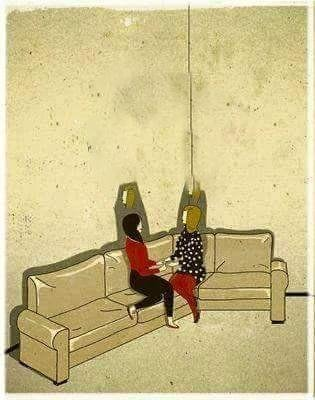}
  \caption{LLM response: no. the image is not inherently funny. it depicts two people sitting on a couch, and there's no clear comedic element. Range rating: 3}
\end{subfigure}

\vspace{2em} 
\begin{subfigure}{0.24\textwidth}
  \includegraphics[width=\linewidth]{figures/gpt5.png}
  \caption{LLM response: no. it's a standard photo of a person speaking.  there is nothing humorous. Range rating: 2}
\end{subfigure}
\hfill
\begin{subfigure}{0.24\textwidth}
  \includegraphics[width=\linewidth]{figures/gpt6.png}
  \caption{LLM response: no. it is a picture of a beach with some people on it. Range rating: 1}
\end{subfigure}
\hfill
\begin{subfigure}{0.24\textwidth}
  \includegraphics[width=\linewidth]{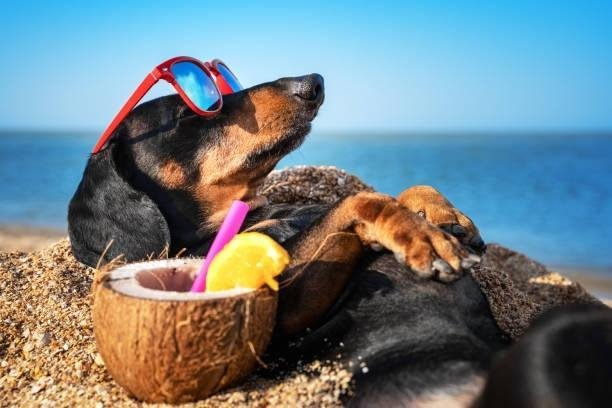}
  \caption{LLM response: yes. the dog is wearing sunglasses and enjoying a coconut drink. Range rating: 8}
\end{subfigure}
\hfill
\begin{subfigure}{0.24\textwidth}
  \includegraphics[width=\linewidth]{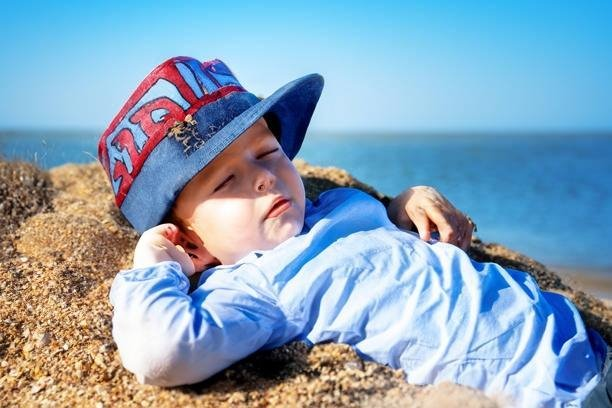}
  \caption{LLM response: no. the image is of a child sleeping.  there is no humor in the image. Range rating: 2}
\end{subfigure}
\caption{\textbf{Example Gemini-Flash binary classification, range ratings, and explanations.} The human binary and range ratings are as follows for the images: (a) \emph{83\%, 6.75}, (b) \emph{14\%, 5.70}, (c) \emph{60\%, 3.60}, (d) \emph{33\%, 3.00}, (e) \emph{60\%, 2.80}, (f) \emph{0\%, 5.57}, (g) \emph{100\%, 5.60}, (h) \emph{33\%, 4.86}.}
\label{fig:gemini_images}
\end{figure*}

\begin{figure*}[h]
\vspace{5pt}
\centering
\begin{subfigure}{0.24\textwidth}
  \includegraphics[width=\linewidth]{figures/gpt7.png}
  \caption{LLM response: Yes. The humor comes from the juxtaposition of a serious medical situation with the casual and nonchalant behavior of the doctors, who seem more interested in using their phones than focusing on the patient. Range rating: 6}
\end{subfigure}
\hfill
\begin{subfigure}{0.24\textwidth}
  \includegraphics[width=\linewidth]{figures/gpt8.png}
  \caption{LLM response: Yes, the image is intended to be humorous. It depicts a cartoon of a patient in a hospital bed with two doctors, one holding a tray of food and the other holding a tray of surgical instruments, both seemingly confused about which tray to use. Range rating: 6}
\end{subfigure}
\hfill
\begin{subfigure}{0.24\textwidth}
  \includegraphics[width=\linewidth]{figures/gemini3.png}
  \caption{LLM response: Yes, the image is funny. It depicts two people sitting on a couch, seemingly engaged in a conversation, but the shadow on the wall behind them forms a shape that looks like a snake eating its own tail. Range rating: 5}
\end{subfigure}
\hfill
\begin{subfigure}{0.24\textwidth}
  \includegraphics[width=\linewidth]{figures/gemini4.png}
  \caption{LLM response: Yes, the image is funny. It depicts two people sitting on a couch, but their bodies are mirrored, with their heads on the wrong sides of their bodies, creating a humorous and surreal visual effect. Range rating: 6}
\end{subfigure}

\vspace{2em} 
\begin{subfigure}{0.24\textwidth}
  \includegraphics[width=\linewidth]{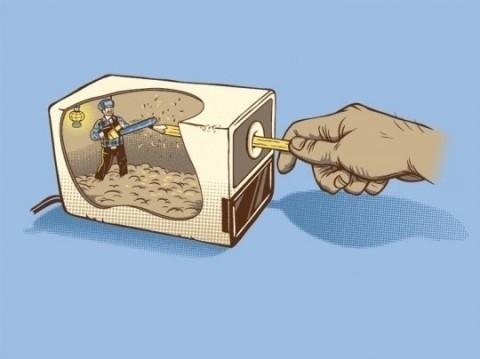}
  \caption{LLM response: Yes, the image is funny. It depicts a cartoon of a person inside a box, seemingly being "pulled out" by a hand using a toothpick. The humor comes from the absurdity and unexpectedness of the situation. Range rating: 6}
\end{subfigure}
\hfill
\begin{subfigure}{0.24\textwidth}
  \includegraphics[width=\linewidth]{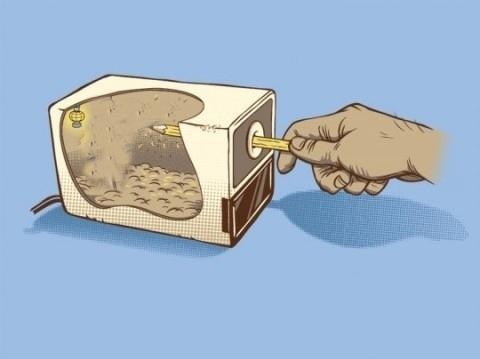}
  \caption{LLM response: Yes, the image is funny because it shows a hand pushing a button on a box that is labeled "Emergency Button," but instead of a typical emergency response, the box is filled with a bunch of pencils, which is an unexpected and humorous outcome. Range rating: 5}
\end{subfigure}
\hfill
\begin{subfigure}{0.24\textwidth}
  \includegraphics[width=\linewidth]{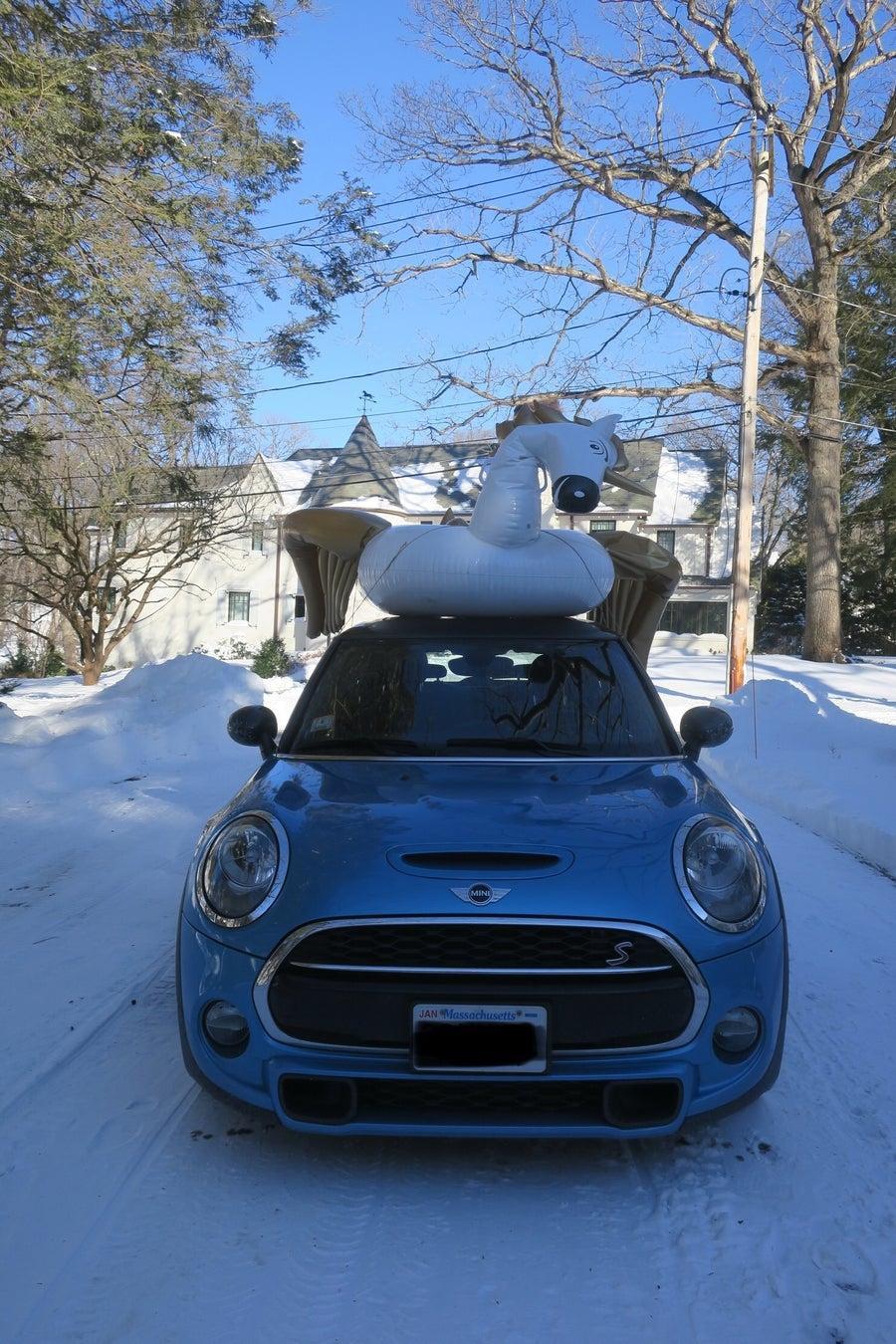}
  \caption{LLM response:Yes, the image is funny because it shows a car with a large inflatable pool toy, specifically a white inflatable animal, strapped to its roof. This is an unusual and humorous sight because it's not common to see such a large and whimsical item being transported in this manner. Range rating: 7}
\end{subfigure}
\hfill
\begin{subfigure}{0.24\textwidth}
  \includegraphics[width=\linewidth]{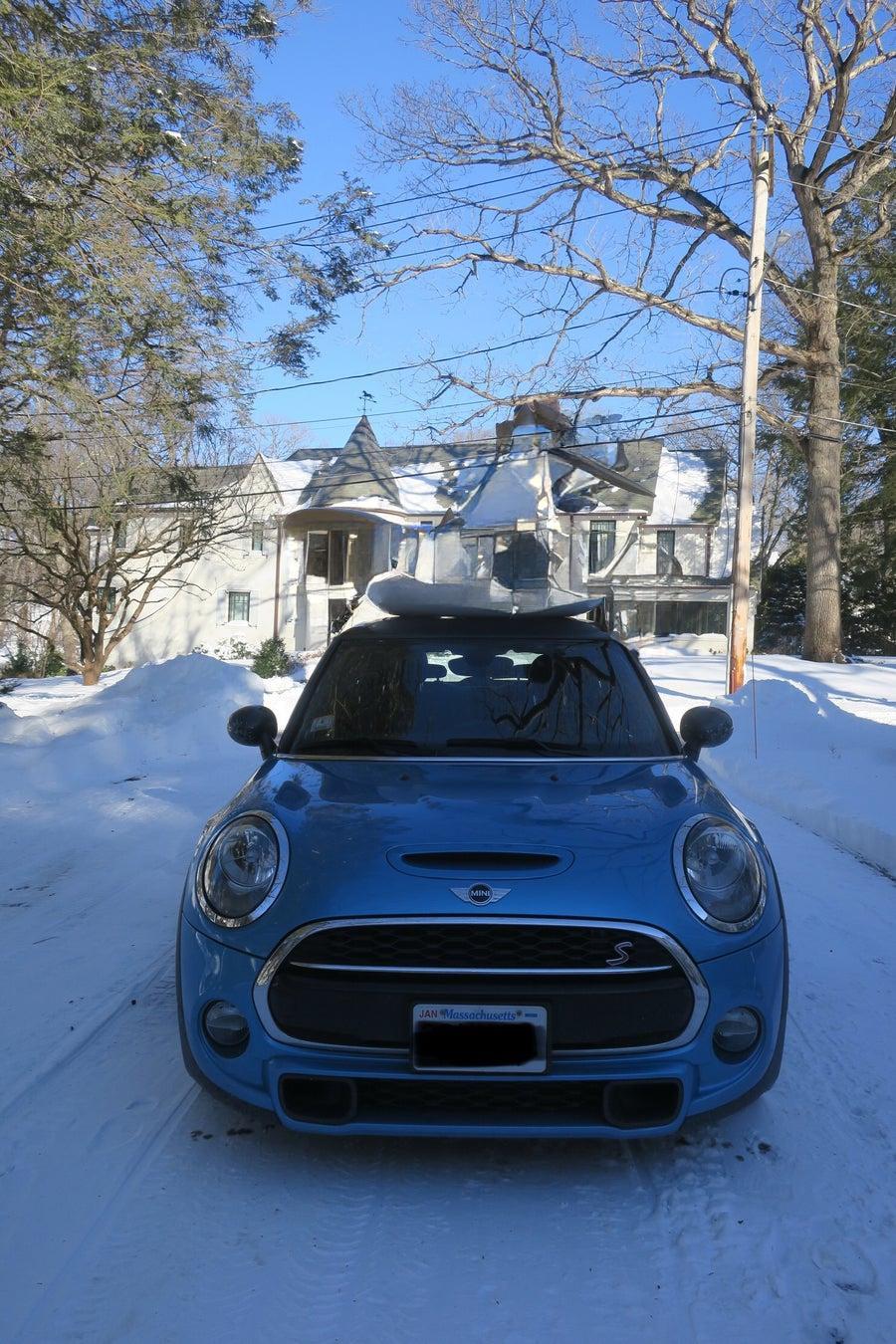}
  \caption{LLM response: No. The image shows a car parked in a snowy environment, which is a common sight during winter. There is nothing inherently funny about the image; it appears to be a typical winter scene. Range rating: 2}
\end{subfigure}
\caption{\textbf{Example LlaVA binary classification, range ratings, and explanations.} The human binary and range ratings are as follows for the images: (a) \emph{83\%, 6.75}, (b) \emph{14\%, 5.70}, (c) \emph{60\%, 3.60}, (d) \emph{33\%, 3.00}, (e) \emph{60\%, 2.80}, (f) \emph{0\%, 3.00}, (g) \emph{50\%, 5.25}, (h) \emph{0\%, 1.00}.}
\label{fig:llava_images}
\end{figure*}

For evaluating zero-shot performance of the large multimodal models we test them on testAllSet and testAllPairs sets. We do this to compare the performance of these models with the other fine tuned models on the same test set. The two variants of prompts in binary task as mentioned in section 4.1 were: : (i) “Is the image funny?”, and (ii) “Is the image not funny?”. The performance on both prompts were similar so we reported the average for the results. In addition for succinct explanations in a particular format for the figures like Fig. ~\ref{fig:gpt_4_images}, we add a suffix 'start answer with yes/no then explain'.
In this section we present similar figures to Fig.~\ref{fig:gpt_4_images} for Gemini-Flash and LlaVA on zero-shot prompting. We also mention the range ratings these models give for the images and the range rating prompt is mentioned in section 4.1.
Gemini-Flash answer explanations are presented in \textbf{Fig.~\ref{fig:gemini_images}}. The
answer explanation from LlaVA zero-shot are presented in \textbf{Fig. ~\ref{fig:llava_images}}.

\newpage

\subsection{Participants' Demographics}

The demographics of the participants are described in the table \textbf{\cref{tab:demographic-data}}.

\label{sec:participant_demo}
\begin{table}[h]
\centering
\begin{tabular}{|l|r|}
\hline
\textbf{Demographic Category} & \textbf{Percentage} \\
\hline
\multicolumn{2}{|l|}{\textbf{Age}} \\
\hline
20-29 & 47.0\% \\
30-39 & 30.3\% \\
40-49 & 13.8\% \\
50-59 & 5.2\% \\
0-19 & 2.1\% \\
60-69 & 1.6\% \\
\hline
\multicolumn{2}{|l|}{\textbf{Education}} \\
\hline
Undergraduate & 56.3\% \\
Postgraduate & 29.5\% \\
High School & 14.2\% \\
\hline
\multicolumn{2}{|l|}{\textbf{Gender}} \\
\hline
Male & 52.6\% \\
Female & 46.6\% \\
Non-binary & 0.8\% \\
\hline
\multicolumn{2}{|l|}{\textbf{Nationality}} \\
\hline
United States & 54.0\% \\
South Africa & 9.0\% \\
Other (41 total) & 37.0\% \\
\hline
\end{tabular}
\caption{Demographic Data}
\label{tab:demographic-data}
\end{table}

\section{Scoring for VLM explanation evaluation}
\label{sec:score_explain}
The scoring process for all words and common words scores involved:
\begin{itemize}
\item Stemming words from human and VLM explanations using the spacy library.
\item Matching synonyms using the nltk library.
\item Marking an explanation as satisfactory if any stemmed word synonym from human raters is contained in the model explanation.
\end{itemize}

\subsection{Results for automated VLM explanation evaluation}
VLM explanations for correctly classified funny images showed varying alignment with human annotations (\textbf{Table \ref{tab:explanation_accuracy}}). Gemini provided the most accurate explanations (87.6\% All Words Score, 74\% Common Words Score), followed by GPT-4o and LLaVA.

\begin{table}[h]
\centering
\begin{tabular}{|l|c|c|}
\hline
Model & All Words Score & Common Words Score \\
\hline
Gemini & 0.876 & 0.74 \\
GPT4o & 0.840 & 0.724 \\
LLaVA & 0.74 & 0.50 \\
\hline
\end{tabular}
\caption{VLM explanation accuracy compared to humans annotations}
\label{tab:explanation_accuracy}
\end{table}

\section{Attention Map Evaluation}
\label{sec:attn_map_eval}
To gain insight into what regions of an image the model relies on to infer humor, we analyzed attention maps in vision transformer models \cite{abnar2020quantifying}. Specifically, we examined the attention on images from the test set. We used an \emph{attention rollout} technique that iteratively multiplies the raw attention matrices (augmented by the identity matrix to include residual connections), producing a global attention map over image tokens. We then reshaped this map into a 2D image by mapping each token to its position in the image grid and upsampling it to the original resolution. Attention Rollout does not work for SwinV2 because of its hierarchical attention architecture.

After computing the final attention map, we optionally convert this attention to a \emph{segmentation-like} mask. We smooth the attention map, detect local peaks, and expand those peaks into connected blobs whose intensities exceed a fraction of the peak value. This yields “peak-centered blobs,” each one capturing a region of high attention. Specifically:

\paragraph{Peak-Centered Blob Detection.}
We developed a systematic approach to identify and analyze significant attention regions:
\begin{enumerate}
    \item Applied Gaussian smoothing ($\sigma$=2.0) to reduce noise in attention maps
    \item Detected local maxima using peak\_local\_max with relative threshold 0.5
    \item Generated connected components around peaks using intensity-based thresholding
    \item Filtered small regions (<50 pixels) to focus on significant attention areas
\end{enumerate}

Since our dataset contains minimally contrastive pairs, one can also approximate the “true” humorous region by taking the pixel-level difference between the original (funny) and modified (not funny) image in each pair.

\paragraph{Difference Map Generation.}
For paired images (funny/not-funny versions), we:
\begin{enumerate}
    \item Computed pixel-wise differences across RGB channels
    \item Applied Gaussian smoothing ($\sigma$=1.0) to the difference map
    \item Identified connected components with significant differences (threshold > 0.1)
    \item Generated binary masks highlighting modified regions
\end{enumerate}

\paragraph{Evaluation Metrics.}
We evaluated attention map quality using three primary metrics:
\begin{itemize}
    \item Recall: Proportion of ground truth funny regions captured by attention
    \item Strict Box Containment: Binary measure of whether attention stays within ground truth regions
    \item Outside Box Ratio: Proportion of attention allocated outside ground truth regions
\end{itemize}

We could only do this for the ViT and DinoV2 models as attention rollout cannot be directly be applied to SwinV2 type models which is the only other vision only transformer.

\section{Logit Attribution Details}
\label{sec:logit_attr}
Transformer-based NLP models can be probed using \emph{logit attribution} to evaluate how each layer's hidden state contributes to the final output. Here, we adapt the same concept to our ViT-based model, treating each ViT block (and the final classifier) as a “layer.”

\paragraph{Layer-wise Analysis.}
For each layer in the transformer models (ViTs and DINOv2):
\begin{enumerate}
    \item Extracted hidden states from each transformer layer
    \item Applied the classification head to each layer's output regarding that layer as the "last layer" before layernorm and classification head.
    \item Computed softmax probabilities for binary humor classification
    \item Measured classification accuracy using each layer's predictions
\end{enumerate}

\section{Future Directions.}
\label{sec:future_dirs}
Beyond expanding the cultural breadth of \textbf{HumorDB}, we envision several promising research avenues:

\begin{itemize}
    \item \textbf{Interpreting Vision Models.} Recent progress in interpreting transformer-based language models can inform the study of multimodal and vision-only architectures. HumorDB’s carefully constructed minimal pairs provide an ideal testbed for \emph{mechanistic interpretability} in vision, offering the kinds of subtle input differences that are otherwise hard to curate for images.

    \item \textbf{Expanded Evaluation Metrics.} Novel benchmarks could explore multi-modal inputs (e.g., text, video, audio) to capture richer humor contexts. This would help evaluate how well models integrate multiple information streams to detect incongruities or comedic timing.

    \item \textbf{Personalized Humor.} Because individual comedic tastes vary, it would be valuable to test models on how well they adapt to personal preferences. Such personalization could move beyond majority voting to reflect diverse humor perceptions.

    \item \textbf{Cultural and Linguistic Diversity.} Truly universal humor comprehension requires sampling across diverse cultural and linguistic backgrounds. Curating a broader spectrum of comedic tropes—slapstick, satire, wordplay, and so on—will challenge models to generalize beyond Western-centric contexts.

\end{itemize}

\section{In-Lab Validation of Crowdsourced Data}
To validate our primary crowdsourced annotations, we conducted a separate in-lab (non-crowdsourced) study. Participants provided ratings for 400 images, yielding $\ge$5 ratings per image. We found high correlation between the in-lab and online data for both the binary task ($\rho=0.78$) and the range task ($\rho=0.72$), which reinforces the reliability of our HumorDB.

\newpage
\section{Image categories results}

The results of models on various image categories are described in \textbf{\cref{tab:binary_types}}.

\begin{table*}
\centering
\begin{tabularx}{\textwidth}{|X|X|X|X|X|X|}
\hline
Model Name & Photos &  Photoshopped  & Sketches  & Cartoons & AI-Gen \\
\hline
dinov2 large& $59 \pm 3$ & $60 \pm 1$ & $47 \pm 1$& $59 \pm 2$&$51 \pm 2$\\
\hline
vit huge& $64 \pm 3$ & $61 \pm 2$ & $48 \pm 2$& $62 \pm 2$&$52 \pm 2$\\
\hline
vit large& $58 \pm 2$ & $58 \pm 2$ & $47 \pm 1$& $59 \pm 2$&$51 \pm 2$\\
\hline
swin2 large & $61 \pm 2$ & $60 \pm 2$ & $47 \pm 1$& $60 \pm 2$&$52 \pm 2$\\
\hline
convnext large& $57 \pm 2$ & $57 \pm 1$ & $46 \pm 1$& $57 \pm 2$&$50 \pm 0$\\
\hline
vitg 14 & $72 \pm 3$ & $70 \pm 3$ & $51 \pm 2$& $68 \pm 3$&$53 \pm 2$\\
\hline
resnet152 & $56 \pm 1$ & $55 \pm 2$ & $46 \pm 1$ & $56 \pm 2$ & $50 \pm 1$ \\ 
\hline
LLaVA (Zero-Shot) & $63 \pm 5$ & $66 \pm 4$ & $46 \pm 1$ & $52 \pm 2$ & $66 \pm 3$ \\
\hline
LLaVA (fine-tuned) & $72 \pm 2$ & $76 \pm 2$ & $54 \pm 2$& $65 \pm 2$&$69 \pm 3$\\
\hline
LLaVA (words fine-tuned) & $79 \pm 2$ & $83 \pm 1$ & $54 \pm 1$& $69 \pm 2$&$73 \pm 1$\\
\hline
BLIP (fine-tuned) & $59 \pm 1$ & $59 \pm 2$ & $48 \pm 1$& $59 \pm 2$&$52 \pm 2$\\
\hline
BLIP (words fine-tuned) & $63 \pm 2$ & $66 \pm 2$ & $49 \pm 1$ & $61 \pm 2$ & $55 \pm 1$ \\
\hline
GPT-4o (Zero-Shot) & $75$ & $69$ & $50$ & $58$ & $76$ \\ 
\hline
Gemini-Flash & $73  $ & $84$ & $53$ & $74$ & $82$ \\
\hline
\end{tabularx}
\caption{Binary Classification Results on various image types present in the dataset.}
\label{tab:binary_types}
\end{table*}

\end{document}